\documentclass{article}
\usepackage{float}
\usepackage{microtype}
\usepackage{graphicx}
\usepackage{subfigure}
\usepackage{booktabs} 
\usepackage{amsmath}
\usepackage{amssymb}
\usepackage{bbold}
\usepackage{multicol}
\usepackage{multirow}
\usepackage{booktabs}
\usepackage{todonotes}

\usepackage{hyperref}
\usepackage{lipsum}

\renewcommand{\paragraph}[1]{\noindent\textbf{#1}}

\usepackage[accepted]{mlsys2025}

\mlsystitlerunning{Dynamic Expert Sparsification for MoEs}

\begin{document}

\twocolumn[
\mlsystitle{KEEP: Dynamic Expert Sparsification \\ for Mixture-of-Experts Models}



\mlsyssetsymbol{equal}{*}

\begin{mlsysauthorlist}
\mlsysauthor{Maximilian Kleinegger}{tuw,ista}
\mlsysauthor{Dan Alistarh}{ista}
\end{mlsysauthorlist}

\mlsysaffiliation{ista}{Institute of Science and Technology Austria (ISTA), Vienna, Austria}
\mlsysaffiliation{tuw}{TU Wien, Vienna, Austria}

\mlsyscorrespondingauthor{Maximilian Kleinegger}{maximilian.kleinegger@ista.ac.at}

\mlsyskeywords{Machine Learning, MLSys}

\vskip 0.3in

\begin{abstract}

Expert pruning, the process of removing a subset of routed experts from a Mixture-of-Experts (MoE) model, is a popular way to reduce the size of MoE models. However, currently, state-of-the-art methods fail to push beyond the 50\% sparsity mark without significant accuracy degradation. Specifically, past this mark, models usually become task-specific or collapse in accuracy.
In this paper, we show that this accuracy degradation is linked to their pruning objective being \emph{static}, that is, independent of the specific user prompt. We find that expert importance is concentrated within individual prompts but varies across prompts, domains, and between prefill and decoding. Motivated by this observation, we propose \emph{KEEP}, which dynamically selects experts for each request. KEEP runs prefill on the full model and uses its routing information with a lightweight, offline-calibrated linear predictor to select the experts retained for decoding. Furthermore, we co-design a serving engine that places the kept set across GPU, host memory and disk, so that decoding never reads from disk. We show that KEEP can successfully push beyond the 50\% sparsity barrier while remaining accurate and being deployable: across three MoEs from 35B to 749B parameters, KEEP retains $\geq$99\% of dense accuracy at 50\% sparsity and 90--97\% at 75\%, outperforming static pruning methods.
Across devices and models, KEEP enables decoding fully on GPU or GPU+host where this was previously not possible, and increases decoding throughput by up to 3$\times$ compared to specialized MoE serving engines such as FreeToken, as well as the widely adopted llama.cpp. Code is available at \url{https://github.com/IST-DASLab/keep-engine}

\end{abstract}
]



\printAffiliationsAndNotice{}  

\section{Introduction}

Recent advances in Large Language Models (LLMs), particularly in coding performance, have increased the popularity of local AI systems~\cite{survey_agentic_ai2, survey_agentic_ai1}, which allow users to run powerful models on inexpensive hardware. 
Yet, executing those models on such devices is still very cumbersome, as strong models exceed hundreds of billions of parameters~\cite{openai2025gptoss120bgptoss20bmodel, qwen38, deepseekai2025deepseekv3technicalreport, Guo_2025, deepseekai2026deepseekv4highlyefficientmilliontoken, kimiteam2026kimik2openagentic, kimiteam2026kimik3openfrontier, glm5team2026glm5vibecodingagentic}. 
This forces the community to either rely on large private vendors' APIs, or invest heavily in cloud or private infrastructure. Consequently, this limits model choice and capabilities, as well as raising privacy concerns. 

The current prevailing model architecture, called Sparsely-Gated Mixture-of-Experts (MoE)~\cite{shazeer2017outrageouslylargeneuralnetworks}, is designed to be very efficient for inference by only activating a fraction of parameters, at the cost of massive memory requirements.
Compression methods address this challenge through \textit{quantization}, e.g.~\cite{gptq, lin2024awq} or various \textit{pruning} techniques~\citep{frantar-sparsegpt, wanda, lasby2026reap}.
However, whether by low-bit quantization or sparsity, pushing towards very low memory footprints induces clear trade-offs between size and accuracy. 

A clear example in the context of expert pruning, which is the focus of our paper, is that of REAP, a current state-of-the-art approach~\cite{lasby2026reap}: when pruning more than 50\% of experts, as required for local deployment of large models, the method only provides reasonable accuracy for \textit{specific domains}, correlated with the calibration data (e.g., \textit{coding} tasks), but clearly drops quality across general tasks (e.g., \emph{knowledge-intensive} tasks). 

In this paper, we take a different approach to MoE pruning. We ask whether MoEs can be \emph{dynamically sparsified} at runtime, \textit{based on their input prompt}, to strictly improve the accuracy-size trade-off, while keeping latency low. From the systems' perspective, we start from the idea of \emph{streaming} model experts directly from RAM or SSD when necessary~\citep{xue2025moeinfinityefficientmoeinference, 10.1145/3669940.3707267, tang2024hobbitmixedprecisionexpert, yang2026freetokenefficientedgenativemoe}, but only selecting the subset that is relevant to the current task.  
Thus, we aim for an algorithm-system co-design that can adapt to any memory budget (e.g. GPU and GPU-plus-host memory), while maximizing accuracy in a dynamic, task-specific manner.

We begin with a thorough investigation of expert activation behaviour across models and domains. We find that models' expert activation patterns are sparse, especially with increasing parameter count: roughly, for any task, the answer is generated by only 50\% of all experts. We introduce KEEP (Keep the Experts) an algorithm-system co-design which adaptively sparsifies experts of a MoE, based on the input prompt, to ensure that memory constraints are met. KEEP can be orthogonally combined with other optimizations, such as  quantization. Our contributions are as follows:

\begin{enumerate}
    \item \textbf{KEEP Algorithm.} We propose an algorithm that dynamically sparsifies MoE decoding based on the expert activation patterns during the prefill stage.
    Precisely, during an offline calibration phase, we capture LLM outputs and the respective expert activation patterns during prefill, as well as the importance of experts during decoding.
    We continue by training a lightweight predictor between the prefill activation patterns and the most important experts during decoding.
    While serving, we capture the expert activation patterns during prefill, which are then used to predict and rank the most important experts for decoding. The remaining experts are ultimately masked out, restricting decoding to our sparsified set.
    
    \item \textbf{KEEP System.} Furthermore, we propose a system accompanying our algorithm. Our system works by tracking the activated experts during prefill, and distributing the retained experts based on importance across GPU and host memory. Via optimized prefill and decoding phases, we enable inference on memory-restricted systems.
    Our system is built on top of FreeToken~\citep{yang2026freetokenefficientedgenativemoe}, allowing to reuse existing optimization techniques such as quantization or more advanced system optimizations. 
    \end{enumerate}

We experimentally validate our design on two recent open model families (Qwen and DeepSeek) across multiple benchmarks covering math, reasoning, knowledge, and coding, as well as agentic tasks for the latest Qwen model. 
Overall, we show that KEEP manages to recover $\geq 99\%$ of the full-model performance at 50\% sparsity across domains and over 90\% of full-model performance at high (75\%) sparsity levels. Furthermore, KEEP enables either decode on GPU+host or full GPU enabling decoding speedups up to $3\times$ with negligible overhead during prefill.

\section{Related Work}

\textbf{Compression.} Quantization and pruning are standard approaches to reduce the memory footprint of LLMs. Widely adopted methods include GPTQ~\cite{gptq} and AWQ~\cite{lin2024awq} for scalar quantization, as well as SparseGPT~\cite{frantar-sparsegpt} and Wanda~\cite{wanda} for unstructured pruning. The compression rates of those methods are often insufficient for memory-constrained deployment.
Stronger quantization can be achieved through more complex methods, such as QuIP\#~\cite{tseng2024quipbetterllmquantization}, AQLM~\cite{AQLM} and QTIP~\cite{Tseng2024_QTIP}. However, those methods also struggle to fully recover model performance at very low bitwidths ($\leq 2$ bits per parameter).

For structured pruning of experts, tailored approaches such as EASY-EP~\cite{dong2025domainspecific}, REAP~\cite{lasby2026reap}, DiEP~\cite{bai2025diep}, EvoESAP~\cite{liu2026evoesapnonuniformexpertpruning} and EvoPess~\cite{sieberling2025evopress} were proposed. However, these methods still struggle to push beyond expert sparsity levels of 50\%: either accuracy drops across domains, or the remaining model tends to perform well only across a specific set of tasks that are correlated with the calibration data. 
Therefore, with current methods, there is a clear accuracy-size trade-off, where we trade model quality for the ability to run models on-device.

Our work does not remove this limitation (which may well be inherent). Instead, it shows that \emph{dynamic input-specific pruning} is a powerful way to bridge the gap between memory costs and accuracy. 

\textbf{``Dynamic'' Pruning.} The above methods are \emph{static}: pruning is performed once and offline. Dynamic methods make sparsity input-dependent. In dense models, Deja Vu~\cite{liu2023dejavu} exploits \emph{contextual sparsity}, predicting per token which attention heads and MLP neurons can be skipped, while \citet{lu-etal-2024-experts} complement static expert pruning with dynamic per-token skipping. Closer to us, IFPruning~\cite{pruning-large-language} selects a sub-network once per prompt: a small LLM encodes the prompt, and a mask predictor selects the FFN dimensions of a dense model. Yet, the predictor needs continued pre-training, which is expensive. For MoEs, PreMoE~\cite{pei2025premoe} ranks experts with domain-aware scores derived from router logits and compiles sparse variants for target domains ahead of time, for a fixed set of deployment scenarios. SlimWise~\cite{park2026slimwisedecouplingexpertpruning} decouples phases, running prefill on the full model and decoding on a pruned model that reuses the full-model KV cache. Yet, the pruned decode model is identical across requests. We use SlimWise plus REAP as our static baseline (Section~\ref{sec:setup}).

\textbf{Concurrent work.} Dynamic Expert Pruning (DEP)~\cite{koo2026dep} transfers the IFPruning predictor to MoEs: a small LLM encodes an agent's system and task prompt, and a shallow MLP head maps it to a per-request expert mask. DEP focuses on accuracy: experiments run with all experts resident, and loading costs are only bounded theoretically. \citet{dettmers2026runtimedynamiccompression} jointly allocates per-layer quantization levels and expert-removal ratios to meet a memory budget, and adapts the resident expert set during decoding by exchanging up to four experts per layer every 256 tokens. However, this approach is evaluated solely in terms of perplexity.

KEEP differs from these approaches in three respects: 1) it doesn't require auxiliary encoders nor backbone modifications, 2) the kept set is computed for each individual prompt, rather than compiled statically and 3) KEEP is co-designed with the inference system: the kept set directly determines expert placement across GPU, host memory and disk.

\textbf{Expert Offloading} is an orthogonal approach, where expert weights are stored in host memory or on disk and transferred to the GPU on demand. Systems such as MoE-Infinity~\cite{xue2025moeinfinityefficientmoeinference}, MoE-Lightning~\cite{10.1145/3669940.3707267} and FreeToken~\cite{yang2026freetokenefficientedgenativemoe} mitigate this overhead via expert caching, prefetching, and scheduling. There is a very large body of work on prefetching strategies, such as DAOP~\cite{zhang2025daop}, CasMoE~\cite{wang2026casmoe}, DoMoE~\cite{mu2026domoe}, Speculating Experts~\cite{madan2026speculating}, and MoE-CORE~\cite{yang2026moecore}. Another line of work reduces the overhead of cache misses~\cite{tang2024hobbitmixedprecisionexpert, wang2025d, zhou2025floe, yang2026zipmoeefficientondevicemoe}. 
KEEP builds on FreeToken and is complementary to most of these techniques. Several of them, e.g., CasMoE and DoMoE, also exploit the prompt, but to steer prefetching and caching: routing is left unchanged. KEEP instead restricts decoding to a prompt-specific kept set sized to the available memory.

\section{Motivation}
\label{sec:analysis}

\begin{figure*}[t]
    \centering
    \subfigure[Saliency concentration]{%
        \includegraphics[width=0.55\textwidth]{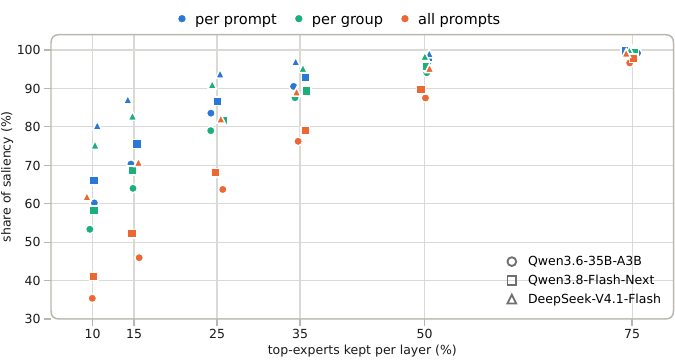}%
        \label{fig:expert_frequency}}
    \hfill
    \subfigure[Expert selection quality]{%
        \includegraphics[width=0.38\textwidth]{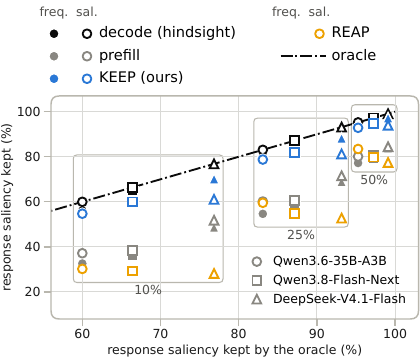}%
        \label{fig:predictor-quality}}

    \caption{\textbf{(a)} Share of saliency captured by the top-\% highest-saliency experts per MoE layer (mean over layers) on 24 prompts from 8 task groups, for 10--75\% of experts kept. Experts are ranked \emph{per prompt} (one ranking per prompt, mean over prompts), \emph{per group} (one ranking per task group, mean over groups), or \emph{across all prompts} (one ranking). Points are jittered horizontally.
    \textbf{(b)} Share of response saliency captured by the top-\% experts per layer, against the oracle (dash-dot), the response's own top-\% by saliency in hindsight. Experts are selected by frequency or saliency from the decode (hindsight), from the prefill, by REAP (one static set), or by KEEP (ours).}
    \label{fig:motivation}
\end{figure*}

\begin{figure}[h]
    \centering
    \includegraphics[width=0.90\columnwidth]{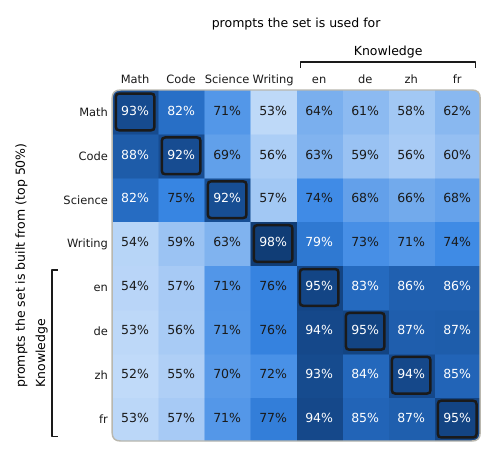}
    \caption{Cross-domain expert saliency for Qwen3.6-35B-A3B. Rows represent the top 50\% highest-saliency experts per layer selected from each task group. Cells show the fraction of column-group saliency captured, averaged over layers and 3 prompts. Boxed cells indicate matching selection and evaluation groups.}
    \label{fig:transfer}
\end{figure}

Existing expert pruning methods consider the pruning objective either within one domain (i.e., math, reasoning, knowledge, and coding) or globally across domains. This means that their calibration data heavily determines the best set of experts. Therefore, we revisit expert contribution patterns across domains and analyze how the most important experts contribute within individual prompts, within groups of prompts, and across multiple domains. Furthermore, we analyze how expert importance changes between the prefill and decoding phases.

We start by revisiting how experts are selected and contribute to the output in MoE models. Each MoE layer contains $K$ experts $f_1, \ldots, f_K$, where $f_k: \mathbb{R}^d \to \mathbb{R}^d$. For each input token $x$, a router selects the top-$k$ experts, denoted by $E(x)$, and assigns them normalized weights $g_j(x)$, while inactive experts receive zero weight. The layer output is then the weighted sum of the selected experts' outputs:

\begin{equation}
h(x) = \sum_{j \in E(x)} g_j(x) f_j(x).
\end{equation}
It has already been shown that expert activation frequency and router weights alone are bad indicators for expert contribution \cite{lasby2026reap, dong2025domainspecific}. 
Based on the observations by \citet{lasby2026reap}, a better criterion is saliency $S_j$. Unlike these simpler metrics, \emph{saliency} also accounts for the magnitude of an expert's actual output, thereby capturing not only how often or how strongly an expert is selected, but also how much it contributes when activated. It is defined as:
\begin{equation}
    S_j = \frac{1}{|\mathcal{X}_j|} \sum_{x \in \mathcal{X}_j} g_j(x) \cdot \|f_j(x)\|_2
    \label{eq:reap_saliency}
\end{equation}
where $\mathcal{X}_j$ is the set of tokens for which expert $j$ is active (i.e., $\mathcal{X}_j = \{x \mid j \in E(x)\}$). The metric is normalized over the tokens to which the expert contributes. Empirically, REAP has shown this criterion to be more effective for expert pruning than frequency- or router-based metrics.

\textbf{Setup.} To evaluate expert contribution patterns, we compute this saliency score for Qwen3.6-35B-A3B~\cite{qwen3.6-35b-a3b}, Qwen3.8-Flash-Next~\cite{qiu2026designqwen38nextarchitectureevaluation}, and DeepSeek-V4.1-Flash~\cite{deepseekai2026deepseekv41flashpushinglimitskv}. We use three prompts per domain, covering math, coding, science, creative writing, and knowledge tasks, across four languages: English, German, Chinese, and French. Further details are provided in Appendix~\ref{sec:expert_pred_ablation}.

\subsection{Expert Usage by Prompt}

We start by analyzing how large the share of total saliency captured by the top-\% highest-saliency experts is, considering individual prompts, groups of prompts within one domain, and prompts across multiple domains. In Figure~\ref{fig:expert_frequency}, we can see how concentrated expert saliency is. Across multiple models, the top-25\% highest-saliency experts capture 84--94\% of the total saliency, while the top-50\% capture 96--99\%. However, when considering the top-25\% highest-saliency experts shared across prompts within one domain, these numbers drop to 79--91\% and 94--98\%, respectively. When considering a single set of highest-saliency experts across all domains, the values decrease further to 64--82\% and 88--95\%. Therefore, the gap widens as more prompts have to share the same set of experts.

\subsection{Expert Saliencies Do Not Transfer Across Domains}

Next, we want to see how well those expert sets, determined by saliency, generalize out-of-distribution. In Figure \ref{fig:transfer}, we can see that the top-50\% experts by saliency, grouped by domain, contribute between 92--98\% within their own domain. However, those generalize poorly out-of-distribution dropping as low as 52\%. Expert sets transfer better between related domains, but still capture less saliency than sets selected directly for the target domain. These results suggest that static expert-sets may fail to capture important experts across diverse domains and should be updated for decoding.

\subsection{Prefill and Decoding Use Different Experts}

\begin{table}[h]
    \caption{Overlap of the top 25\% and 50\% highest-saliency experts between prefill and decoding, averaged over MoE layers and 24 prompts.}
    \label{tab:overlap}
    \vskip 0.15in
    \begin{center}
        \begin{small}
            \begin{tabular}{lcc}
                \toprule
                Model & Top 25\% & Top 50\% \\
                \midrule
                Qwen3.6-35B-A3B     & 59\% & 72\% \\
                Qwen3.8-Flash-Next  & 57\% & 72\% \\
                DeepSeek-V4.1-Flash & 58\% & 69\% \\
                \bottomrule
            \end{tabular}
        \end{small}
    \end{center}
    \vskip -0.1in
\end{table}

Lastly, we consider the difference between the most salient experts during prefill and decoding. In Table~\ref{tab:overlap}, we can see that the expert sets overlap by only 57--59\% for the top-25\% most salient experts, while for the top-50\%, the overlap increases to 69--72\%. This is a substantial difference between the most salient experts during prefill and decoding. Additionally, in Figure~\ref{fig:predictor-quality}, we can see that retaining the most salient experts by prefill do not necessarily recover decoding saliency. However, as an outlook it does not rule out using prefill statistics to estimate the most important decoding experts as indicated by the \emph{KEEP dots}. More details can be found in Appendix~\ref{sec:expert_pred_ablation}.

\begin{figure*}[t]
    \centering
    \includegraphics[width=1\textwidth]{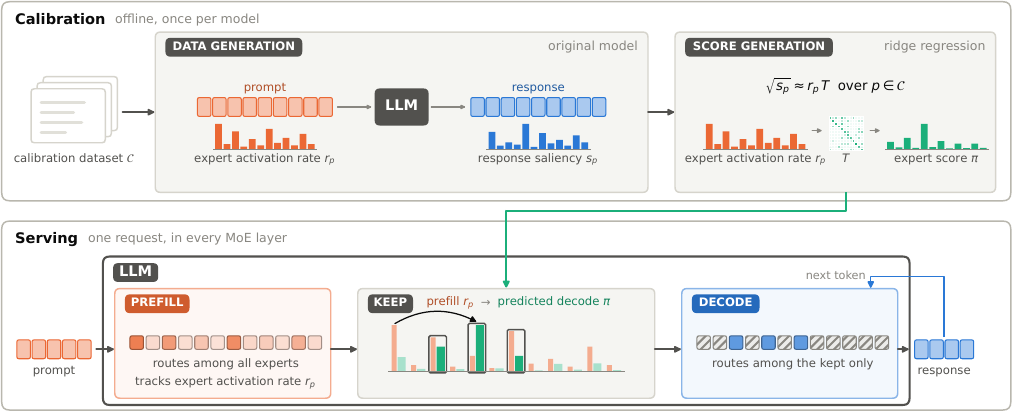}
    \caption{Overview of the method. \emph{Calibration}: The full model processes calibration prompts and responses to train a ridge regression model predicting decoding expert importance from prompt routing patterns. \emph{Serving}: Each prompt is prefilled using all experts. After prefill, each MoE layer retains the top-\% experts by predicted importance (green) for decoding.}
    \label{fig:method}
\end{figure*}

\textbf{Summary.} Expert saliency is concentrated within individual prompts but varies across prompts and domains. This suggests that selecting expert sets for individual requests could preserve more saliency than using a single static set. However, the most important experts during prefill and decoding differ substantially, and decoding saliency is only known after generating the response. Furthermore, experts selected during prefill do not necessarily capture the important experts during decoding.

Given these observations, we focus on sparsifying decoding while keeping prefill unrestricted. The two phases have different memory access patterns: prefill processes prompt tokens in parallel, allowing each expert to be loaded once and shared across tokens. In contrast, decoding generates one token at a time, requiring repeated access to expert weights, which becomes expensive when experts need to be fetched from slower memory. Therefore, sparsifying decoding can avoid these repeated transfers, while pruning prefill would only save its one-time loading cost. Furthermore, prefill benefits from using the full model, as restricting it to only the experts needed for decoding degrades accuracy, especially at high sparsity (Appendix~\ref{sec:pruned_prefill}).

This leaves two challenges for practical deployment: how can we identify the important decoding experts before generation begins, and how can we efficiently maintain request-specific expert sets under memory constraints? We address both challenges in the next section.

\section{Method}
\label{sec:method}

These observations motivate KEEP, an algorithm--system co-design for \textit{dynamic}, request-specific expert sparsification. As illustrated in Figure~\ref{fig:method}, KEEP first uses an \textit{offline} calibration phase to learn how to predict decoding expert importance from prefill routing statistics.
Second, during serving, we run full-model prefill, use the observed activation rates to select the most important decoding experts, and place the retained set across GPU and host memory. We deliberately keep prefill unrestricted to preserve the original prompt processing and collect the routing statistics used by our predictor. Sparsification is applied only to decoding, where the smaller expert working set can reduce memory movement and avoid disk accesses. 

\subsection{Calibration}

\textbf{Data Generation.} During calibration, we sample a set of prompts from the calibration data and generate responses using the model's default generation settings, including sampling and thinking configurations. A teacher-forced replay of each prompt and response then tracks, for every token and expert, whether the expert is activated, its router weight $g$, and the norm of the expert's output $\|f\|$.

\textbf{Score Generation} continues by using these statistics to define the expert activation rate $r_{p,j}$ for a prompt as

\begin{equation}
    r_{p,j} = \frac{|X_j^{p}|}{k \cdot |p|} \quad,
\end{equation}

where $X_j^{p} = \{x \in p \mid j \in E(x)\}$ is the set of tokens for which expert $j$ is active for prompt $p$. We normalize this count by the prompt length and number of activated experts per layer $k$. This normalization ensures that the routing frequencies sum to one across experts within each layer. Although the set of experts selected purely on prefill saliency performs better than the set based on frequency, this advantage disappears, when using it as an input feature for our predictor as shown in Figure\ref{fig:predictor-quality} (more details in Appendix~\ref{sec:expert_pred_ablation}). Therefore, we continue with frequency as it is cheaper to track.

Next, we adapt saliency as proposed by REAP in Equation~\ref{eq:reap_saliency}. We first compute the unnormalized saliency $\tilde{S}_{p,j}$ as follows:
\begin{equation}
    \tilde{S}_{p,j} = \sum_{x \in \mathcal{X}_j^{y_p}} g_j(x) \cdot \|f_j(x)\|_2,
\end{equation}
where $\mathcal{X}_j^{y_p} = \{x \in y_p \mid j \in E(x)\}$ is the set of tokens in response $y_p$ for which expert $j$ is active. Like REAP, we also normalize our saliency:
\begin{equation}
    \label{eq:apex_saliency}
    s_{p,j} = \frac{\tilde{S}_{p,j}}{\sum_{j'} \tilde{S}_{p,j'}}.
\end{equation}
However, instead of normalizing by the number of tokens routed to each expert, as in Equation~\ref{eq:reap_saliency}, we normalize across all experts within a response. Therefore, the saliency scores sum to 1 for each response, giving a normalized prediction target. This captures the relative saliency of each expert for a prompt, rather than its importance at a global level.

\begin{figure*}[t]
    \centering
    \includegraphics[width=1\textwidth]{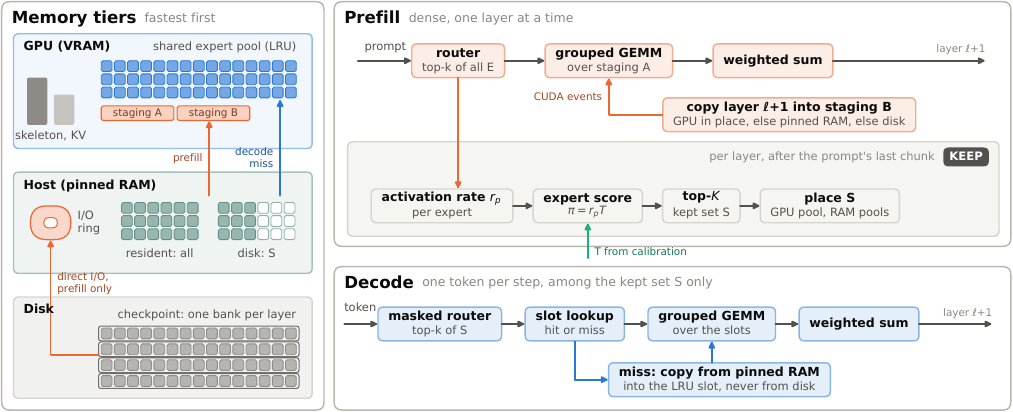}
    \caption{Overview of the KEEP serving engine. Experts are spread over GPU, host RAM and disk (left). Prefill streams every layer and selects the kept set $S$ (top). Decode routes only among $S$ and never reads from disk (bottom).}
    \label{fig:inference}
\end{figure*}

\textbf{Predictor.} Lastly, we learn a simple mapping from the prefill expert activation rate $r_p$ to the square-root-transformed decoding saliency score $s_p$. This allows us to predict which experts will be important during decoding based on their activation during prefill. With this, we account for the differences between the expert sets activated during prefill and decoding.
For each MoE layer, we learn a matrix $T \in \mathbb{R}^{K \times K}$
using ridge regression on the calibration data $\mathcal{C}$:
\begin{equation}
\min_{T} \; \sum_{p \in \mathcal{C}} \big\lVert \sqrt{s_p} - r_p T \big\rVert_2^2 \;+\; \lambda \, \lVert T \rVert_F^2
\end{equation}

We apply a square-root transformation to our saliency scores, to reduce skewness and therefore to reduce the influence of large scores during regression. Since the square root is strictly increasing, this transformation preserves the ranking of the original saliency scores.

Therefore, with KEEP sparsification happens exactly where it needs to be: after the model has processed the prompt, which makes it natively input-aware, but before decoding begins. Since the kept set only depends on the prompt, it is fixed before the first token is generated, which, as we show next, makes KEEP inference-friendly.

\subsection{KEEP Aware Inference}

The second part, as shown in Figure~\ref{fig:method}, is serving with KEEP. The major difference compared to normal serving is the expert activation tracking during prefill, the selection of the most important experts, and the masking of the router during decoding. Therefore, KEEP restricts the experts used during decoding, which enables the following: If the model does not fully fit on the GPU or in GPU+host memory, decoding can run at the full speed of either configuration while relying on streaming from disk during prefill. This either increases decoding speed or enables fast decoding in the first place.

To minimize the runtime overhead of KEEP, we propose 1) a memory-tiered system to enable running models larger than the available memory capacity, 2) optimizing prefill to overlap computation with expert streaming while performing expert selection without any overhead, and 3) designing our expert management as an LRU cache to leverage the high expert concentration.

\subsubsection{Memory Tiers}

Similar to existing LLM and MoE offloading systems~\cite{xue2025moeinfinityefficientmoeinference,yang2026freetokenefficientedgenativemoe,3618408.3619696}, we distribute experts across different memory tiers to reduce GPU memory requirements. We define them by bandwidth as shown in Figure~\ref{fig:inference}: 1) GPU-VRAM, 2) pinned host RAM, and 3) disk memory.

\textbf{Model Distribution.} Following FreeToken~\cite{yang2026freetokenefficientedgenativemoe}, we distribute our model as follows within our system: The model skeleton, therefore everything except the routed experts, and the KV cache are placed on GPU. All experts are placed in pinned host RAM if memory allows, otherwise, we distribute experts between host RAM and disk. The remaining GPU memory is used as a shared expert pool, across layers for decoding.

\textbf{Memory Management.} GPU and host memory are two budgets with three consumers each. On the GPU, the skeleton, the KV cache and the expert pools share the VRAM. Since the skeleton is fixed, a longer context leaves smaller expert pools. On the host, we either store all experts or split the pinned RAM between a pool of resident layers, a pool of disk-backed layers and the I/O ring (a small set of buffers used for reading experts from disk). Since the I/O ring is fixed as well, a lower expert sparsity ratio requires larger pools and therefore moves more layers to disk. Generally, both budgets are determined at startup.

\subsubsection{Prefill Co-Design}

Next, we move to prefill, as illustrated at the top of Figure~\ref{fig:inference}. Prefill in general works as usual and stays dense: the router computes the top-$k$ experts for every token, we perform a grouped GEMM over the selected experts of that layer, and combine the results by their weighted sum, before moving to the next layer. We optimize it in the following: 

\textbf{Double-buffered Prefill.} Following~\cite{yang2026freetokenefficientedgenativemoe,panferov2026disaggregatedquantizationspecializingllm},
we use two staging buffers during prefill, each capable of holding
all experts of one MoE layer. While performing the grouped GEMM
on the experts of layer $\ell$ in one buffer, we load the experts
of layer $\ell+1$ into the other, overlapping computation with
expert loading.
Each expert is fetched from the fastest tier that holds it: experts that already reside within GPU are read in place, experts of host layers and in the pools of disk layers are copied from pinned host memory, and only the remaining experts of disk layers are read from disk through the direct-I/O ring. Therefore, only experts neither in GPU nor in pinned host memory are read from disk, which only happens during prefill. Consequently, disk I/O can become the dominant contributor to prefill latency.

\textbf{Expert Selection.} While prefill runs, we count for every layer how often the prompt routes to each expert, giving us the expert activation rate $r$.
Long inputs are prefilled in chunks~\cite{3691938.3691945}, so these counts accumulate over all chunks. Once the last chunk passes a layer, we compute the importance of each expert for decoding by multiplying the activation rate with our calibrated matrix $T$ as $rT$, and retain the highest ranked expert set $S$ based on our sparsity level.
Then, we place $S$ across our memory tiers:
After processing a layer's final prefill chunk, the top experts from $S$ are copied from the staging buffer, which holds the entire layer at this point, into the shared GPU expert pool, with the available slots distributed uniformly across layers. For host memory layers, nothing else happens and for disk layers, all experts of $S$ are copied into the layer's pinned pool in host memory. Nothing is written back to a lower tier, as host memory and disk always keep their copy. All these operations overlap with the computation of subsequent layers.

\subsubsection{Decode Co-Design}

Decoding, as illustrated at the bottom of Figure~\ref{fig:inference}, works as usual: the router computes the top-$k$ experts over a masked router for every token, we perform a grouped GEMM over the selected experts, and combine the results by their weighted sum, before moving to the next layer. The advantage of our design and the sparsification is that the retained decoding expert set, at the right sparsity level, now fits fully on GPU or on GPU+host, which wasn't possible before.

\textbf{From Expert Pools to LRU.} Similar to FreeToken~\cite{yang2026freetokenefficientedgenativemoe}, we use an LRU expert cache on GPU, where missing experts are copied from pinned host memory into an available slot. If necessary, the least recently used expert with a pinned host copy is evicted, while experts without a pinned host copy are protected from eviction. Since every expert in $S$ remains available either on GPU or in pinned host memory, decoding never requires disk accesses, even when the full model exceeds available host memory.

\subsection{Hardware-Efficient Implementation}

We build upon FreeToken~\cite{yang2026freetokenefficientedgenativemoe}, an inference engine designed for MoEs. We choose FreeToken, as they provide an efficient FTW weight format, which stores the experts of each layer as contiguous, aligned banks that can be pinned in host memory and read from disk with direct I/O. Furthermore, we reuse its LRU expert cache, its double-buffered prefill, its CUDA-graph decode and its serving stack including prefix caching.

On top of it, we add the tracking of expert activations and the selection of the kept set during prefill, the placement of the kept experts across the memory tiers, the restriction of decode to the kept set, and a disk tier for models that exceed host memory. Furthermore, prefill reads experts already cached on the GPU in place, and prefix caching restores the expert activations of a cached prefix, so the kept set does not change. The GPU-side decoding operations are captured in CUDA graphs, while host-to-device transfers and cache management are handled separately.

\textbf{Overlapped three-tier prefill.} Even though FreeToken already supports double-buffered prefill, they don't support streaming from disk, as needed in our three-tier memory layout. FreeToken implements the prefill for GPU+host with two streams. First, there is a copy stream that fills the staging buffers by copying the experts from pinned host memory, and concurrently a compute stream that runs the grouped GEMMs. We extend it by a reader thread that streams disk layers with direct I/O through a small ring of pinned buffers. Since our prefill is bound by the memory bandwidth of the slowest tier, we want to minimize the time the GPU waits for a disk layer. We optimize those disk reads by making them partial: only experts that are neither on GPU nor in pinned memory are read, while the remaining parts of the buffer are fetched from pinned memory and the GPU.
Furthermore, we distribute the layers across the model depth and memory tiers. Therefore, instead of all disk layers being at the head or tail of the model, we keep roughly every $n$-th layer on disk, depending on our memory budget. Lastly, the staging buffers reside within the expert pool, so during prefill, we evict the decoding experts occupying these slots to make room for prefill experts.

\textbf{Selection and placement during prefill.} Expert selection runs entirely on the GPU: the activations are counted per layer and the kept set is ranked by $rT$ once the last chunk passes a layer, without a single host synchronization. The placement of expert set $S$ is planned by a single Triton~\cite{triton} kernel, which chooses the target slots for the most important experts on GPU, which then are executed as a device-to-device copy from the staging buffer. The placement runs in a separate copy stream ordered by CUDA events against the double buffer, therefore overlapping with the computation of following layers. Additionally, for layers residing on disk, the experts of $S$ are copied from the staging buffer into the pinned host memory pool. Furthermore, we store the activated experts of every token with the KV-Cache so that we can enable prefix caching.

\textbf{Graph-safe decode.} 
The GPU-side decoding operations, including cache management and expert loading, are captured in CUDA graphs. Host-driven data transfers and disk I/O are handled separately. Even though the mask changes with every request, we define it as a static tensor, which can be updated in place. Hits, misses and LRU victims are all tracked and resolved within the GPU, and one copy kernel reads missing experts directly from pinned host memory. Lastly, during decoding we also record the unmasked router's top-k selection for every generated token, so that on prefix reuse the expert activation tracker restores the full-model routing distribution. The cached KV entries, however, are heterogeneous: prompt tokens were computed with the full expert set, generated tokens with experts restricted to $S$.

\section{Experiments}

\subsection{General Setup}
\label{sec:setup}

Moving forward, we report our results in terms of \emph{expert sparsity}, rather than the fraction of experts retained as in Section~\ref{sec:analysis}. We define expert sparsity as the fraction of experts excluded from decoding in each MoE layer. Thus, 75\% sparsity corresponds to retaining 25\% of experts.

\textbf{Models \& Data.} We evaluate our method on Qwen3.6-35B-A3B~\cite{qwen3.6-35b-a3b}, Qwen3.8-Flash-Next~\cite{qiu2026designqwen38nextarchitectureevaluation}  and Deepseek-V4.1-Flash~\cite{deepseekai2026deepseekv41flashpushinglimitskv}. Since our method is orthogonal to quantization, we also use quantized models for evaluation. As we see in Table~\ref{tab:models}, those models differ in size, number of total experts, number of activated experts and are increasingly more sparse. To establish a point of comparison, we independently prune those models using the state-of-the-art expert pruning method, REAP~\cite{lasby2026reap}, as well as using the SlimWise~\cite{park2026slimwisedecouplingexpertpruning} technique, which couples the full model for prefill, with our independently pruned REAP models for decode. Therefore, SlimWise is prior work, which uses one static pruned model for decoding, rather than being prompt-aware. We describe the calibration data and composition in Appendix~\ref{section:calibration_data}.

\textbf{Evaluation.} Our goal is to evaluate all models on a diverse set of benchmarks including math, reasoning, knowledge and coding. Therefore, we use the following benchmarks: Math500~\cite{hendrycks2021measuringmathematicalproblemsolving}, GPQA-Diamond~\cite{rein2024gpqa}, MMLU-Pro~\cite{wang2024mmluprorobustchallengingmultitask}, MBPP+~\cite{austin2021programsynthesislargelanguage, evalplus}, Humaneval+~\cite{chen2021codex, evalplus}. Furthermore, we evaluate Qwen3.8-Flash-Next on agentic coding and tool-calling using SWE-Bench Verified~\cite{jimenez2024swebench} (multi-turn) and BFCL-v4~\cite{patil2025bfcl}, respectively. All models used thinking during evaluation, with Qwen3.8-Flash-Next and DeepSeek-V4.1-Flash using their highest settings, \texttt{xhigh} and \texttt{max}, respectively. Thinking was disabled for MMLU-Pro due to compute constraints. Full hyperparameters are provided in Appendix~\ref{sec:exp_design}.

\textbf{Hardware.} Calibration and evaluation were performed on an NVIDIA B200. To accelerate evaluation, we adapted KEEP for batched inference by computing per-request expert activation rates and applying request-specific decoding masks. End-to-end experiments ran on two machines: a desktop with an RTX 5090 and a server with an RTX 3090. Hardware specifications and memory capacities are listed in Table~\ref{tab:system}.

\subsection{Accuracy Results}

\begin{figure*}
    \centering
    \includegraphics[width=0.93\linewidth]{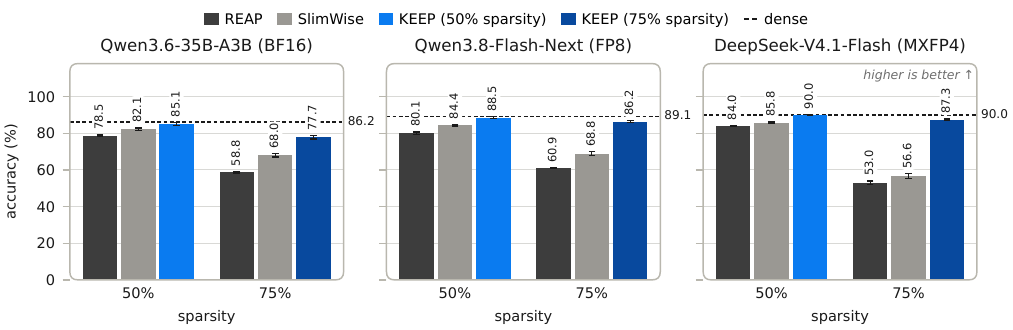}
    \caption{Mean accuracy over MATH-500, GPQA-Diamond, MMLU-Pro, HumanEval+ and MBPP+ at 50\% and 75\% sparsity, for KEEP, REAP and
    SlimWise, with the dense model dashed. Bars show the mean over three sampling seeds}
    \label{fig:results}
\end{figure*}

\begin{figure}
    \centering
    \includegraphics[width=0.95\linewidth]{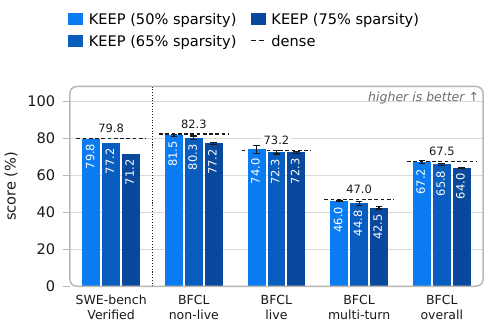}
    \caption{SWE-bench Verified and BFCL-v4 for Qwen3.8-Flash-Next under KEEP at 50--75\% sparsity, with the dense model dashed}
    \label{fig:agentic}
\end{figure}

\paragraph{Main Results.} Figure~\ref{fig:results} shows us the average results across three seeds with standard deviation. Across all three models KEEP manages to recover $\geq 99\%$ of model performance at 50\% sparsity. At 75\% sparsity KEEP drops from 85.4 to 76.7 for Qwen3.6-35B-A3, while managing to stay within 2--3 points for Qwen3.8-Flash-Next and DeepSeek-V4.1-Flash. The models pruned with REAP drop around 6-8 points on average at 50\% sparsity, while at 75\% sparsity they drop around 26--29 points for Qwen models and nearly 40 points for DeepSeek-V4.1-Flash. Therefore, already falling behind at 50\%, and being substantially degraded at 75\% compared to dense. On the other hand SlimWise manages to recover a few more points to compared with standalone REAP. At 50\% sparsity it loses only 3–-5 points compared to dense, while improving 1--4 points over REAP. However, even though it recovers up to 10 points more than REAP at 75\% sparsity, it is still considered substantially degraded. 

Considering all results individually, as seen in Table~\ref{tab:full} (in Appendix~\ref{sec:appendix_results}), we see the biggest drops in accuracy in MMLU-Pro and GPQA-Diamond. Therefore, we can assume that experts, important for math and coding, are overrepresented, even though our calibration mix is multi-domain. These results highlight the difficulty of maintaining accuracy across diverse tasks using a single statically selected expert set. However, the stronger recovery of SlimWise shows us that higher prefill-quality improves recovery.

\textbf{Comparison to Concurrent Work.} Since neither DEP~\cite{koo2026dep} nor Dynamic Runtime Compression~\cite{dettmers2026runtimedynamiccompression} has released its source code at the time of writing, we cannot directly compare against these methods. However, both approaches use pruned prefill, whose impact we analyze in Appendix~\ref{sec:pruned_prefill}. Our analysis suggests that pruned prefill either degrades model performance or provides no memory benefit.

\begin{table}[t]
\centering
\small
\setlength{\tabcolsep}{2pt}
\caption{Generated tokens (millions), with percentage change relative to Dense.}
\label{tab:tokens}
\begin{tabular}{@{}lccc@{}}
\toprule
 & Qwen3.6 & Qwen3.8 & DeepSeek-V4.1 \\
\midrule
Dense & 79\,{\tiny(0\%)} & 66\,{\tiny(0\%)} & 22\,{\tiny(0\%)} \\
\midrule
\multicolumn{4}{@{}l}{\textit{50\% sparsity}} \\
\quad REAP & 109\,{\tiny($\blacktriangle$38\%)} & 94\,{\tiny($\blacktriangle$42\%)} & 58\,{\tiny($\blacktriangle$164\%)} \\
\quad SlimWise & 105\,{\tiny($\blacktriangle$33\%)} & 84\,{\tiny($\blacktriangle$27\%)} & 53\,{\tiny($\blacktriangle$141\%)} \\
\quad KEEP& \textbf{86}\,{\tiny($\blacktriangle$9\%)} & \textbf{66}\,{\tiny(0\%)} & \textbf{23}\,{\tiny($\blacktriangle$5\%)} \\
\midrule
\multicolumn{4}{@{}l}{\textit{75\% sparsity}} \\
\quad REAP & 190\,{\tiny($\blacktriangle$141\%)} & 117\,{\tiny($\blacktriangle$77\%)} & 437\,{\tiny($\blacktriangle$1886\%)} \\
\quad SlimWise & 174\,{\tiny($\blacktriangle$120\%)} & 105\,{\tiny($\blacktriangle$59\%)} & 377\,{\tiny($\blacktriangle$1614\%)} \\
\quad KEEP & \textbf{131}\,{\tiny($\blacktriangle$66\%)} & \textbf{93}\,{\tiny($\blacktriangle$41\%)} & \textbf{26}\,{\tiny($\blacktriangle$18\%)} \\
\bottomrule
\end{tabular}
\end{table}

\begin{figure*}[t]
    \centering
    \includegraphics[width=0.93\textwidth]{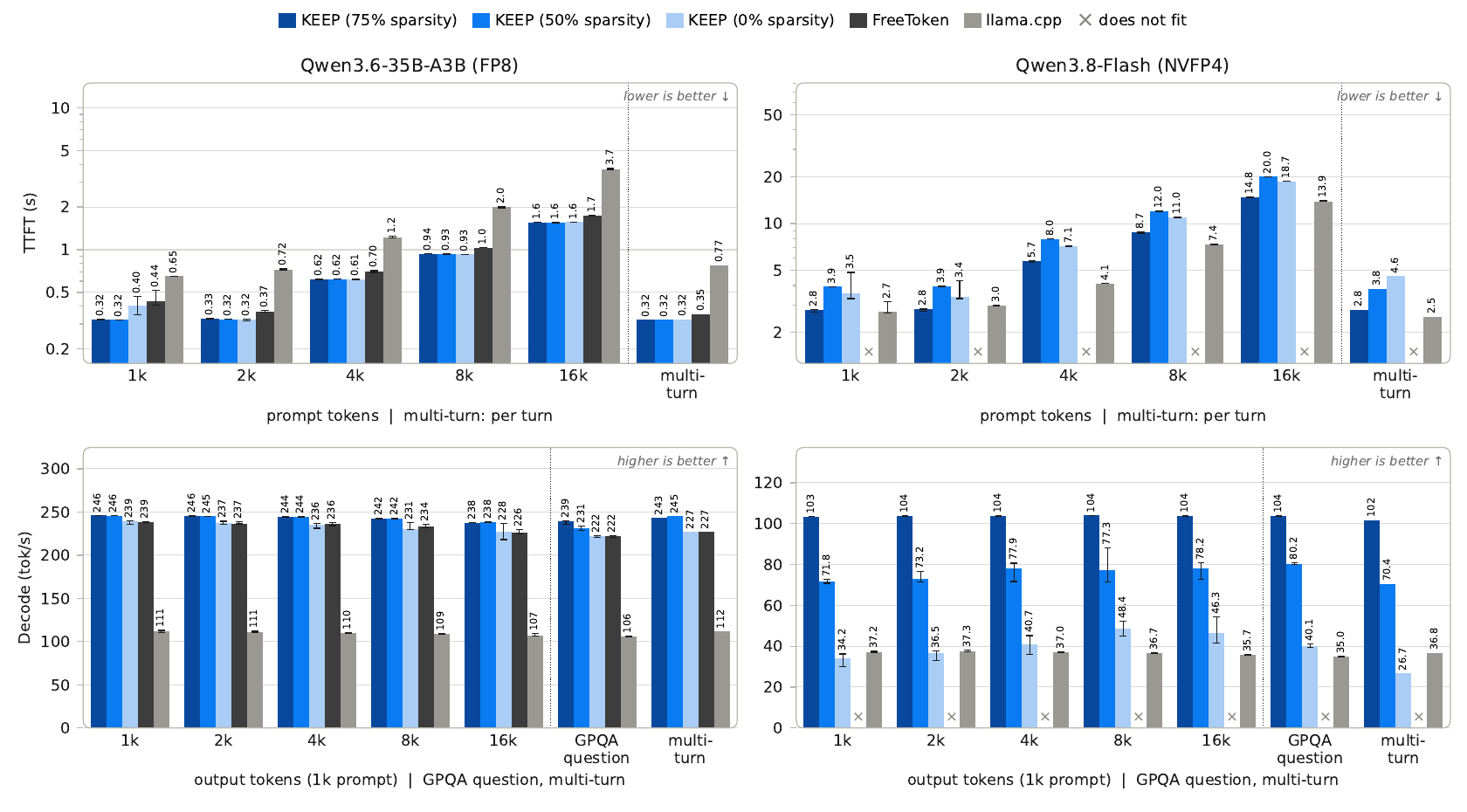}
    \caption{End-to-end evaluation on a desktop RTX 5090. TTFT by prompt length (top) and decode throughput by output length (bottom), and the multi-turn task as the last groups. Bars are medians over four runs after a warm-up. $\times$ means the model does not fit our system.}
    \label{fig:e2e_5090}
\end{figure*}

\textbf{Token Budget.} Lastly, we compare the total number of generated tokens across benchmarks (Table~\ref{tab:tokens}). At 50\% sparsity, KEEP increases token usage by at most 9\%, compared to 30--140+\% for other methods. At 75\%, KEEP still uses the fewest tokens, increasing usage by 18--66\%, compared to 60--1800\% for other methods. These increases can be attributed to model degradation, such as excessive thinking or repetitive loops.

We conclude that KEEP is superior to existing static pruning methods and manages to generalize out-of-distribution, further pushing the accuracy-sparsity Pareto frontier. 


\subsection{Dynamic Sparsification on Agentic Tasks}

Since KEEP enables and accelerates local AI inference, we also wanted to evaluate it on agentic use cases. We run Qwen3.8-Flash-Next on SWE-Bench Verified and BFCL-v4 to evaluate agentic coding and tool-calling. Due to the high evaluation cost, we only evaluate KEEP to measure potential degradation. As shown in Figure~\ref{fig:agentic}, KEEP at 50\% sparsity is lossless on both benchmarks, while at 65\% it loses only 2 points. At 75\%, we see larger drops from 79.8 to 71.2 on SWE-Bench Verified and from 67.5 to 64.0 on BFCL-v4. Therefore, KEEP enables up to 65\% sparsity with minimal loss on real agentic coding and tool-calling tasks.

\subsection{End-to-End Performance}

Lastly, we compare our KEEP-Engine against FreeToken~\cite{yang2026freetokenefficientedgenativemoe} and the widely adopted llama.cpp~\cite{gerganov_llamacpp_2026} to show that our expert selection introduces negligible prefill overhead while improving decoding. We use compressed model versions that fit our machines. The benchmark is described in Appendix~\ref{sec:hw_bench}

In Figure~\ref{fig:e2e_5090}, we compare time-to-first-token (TTFT) and decoding throughput across different prompt and output lengths, as well as multi-turn conversations to evaluate prefix caching. For Qwen3.6-35B-A3B (FP8), KEEP introduces no additional prefill overhead and even outperforms llama.cpp. In decoding, we clearly outperform llama.cpp and stay slightly ahead of FreeToken at 25--50\% sparsity. Since the model is relatively small, KEEP fits the retained experts fully on GPU, while FreeToken relies on GPU+host memory, limiting our advantage.

For the larger Qwen3.8-Flash-Next (NVFP4), which cannot run with FreeToken due to insufficient GPU+host memory, KEEP enables inference. Compared to llama.cpp, TTFT is slightly slower above 50\% sparsity, but decoding throughput is 2--3$\times$ higher. At 75\% sparsity, the retained experts fit fully on GPU, while at 50\% we rely on our LRU cache.

As a reference, KEEP at zero sparsity performs similarly to FreeToken, confirming negligible prefill overhead. Additional results, including further performance metrics and experiments on our RTX 3090, can be found in Appendix~\ref{sec:appendix_e2e}.

\section{Discussion}

In this work, we push beyond the 50\% expert sparsity level of MoEs by introducing KEEP. By determining which experts will be needed for decode, it achieves up to 75\% sparsity while retaining up to 90--97\% of the original model's quality, as demonstrated on three models from the Qwen and DeepSeek family varying size, setting a new Pareto Frontier for pruning, when compared to state-of-the-art pruning methods such as REAP or SlimWise. Furthermore, an optimized and efficient KEEP engine based on FreeToken, shows that determining an expert set for decoding during prefill introduces negligible overhead, while either enabling decoding on device or accelerating it up to 2--3x.

\section*{Acknowledgements}
We thank Andrei Panferov for his valuable feedback and helpful suggestions on the manuscript.

\bibliography{references}

@article{survey_agentic_ai1,
	author = {Bajoria, Sparsh and Ranjan, Shreyanshu and M, Adhitya and Suri, Vineet and Hassija, Vikas and Chamola, Vinay and Hussain, Amir},
	date = {2026/08/24},
	doi = {10.1007/s12559-026-10619-1},
	id = {Bajoria2026},
	isbn = {1866-9964},
	journal = {Cognitive Computation},
	number = {1},
	pages = {103},
	title = {From Language Models to Agentic AI: A Survey of Autonomous, Action-Enabled, and Collaborative LLM Agents},
	url = {https://doi.org/10.1007/s12559-026-10619-1},
	volume = {18},
	year = {2026}}

@article{survey_agentic_ai2,
    author = {Jiang, Juyong and Wang, Fan and Shen, Jiasi and Kim, Sungju and Kim, Sunghun},
    title = {A Survey on Large Language Models for Code Generation},
    year = {2026},
    issue_date = {February 2026},
    publisher = {Association for Computing Machinery},
    address = {New York, NY, USA},
    volume = {35},
    number = {2},
    issn = {1049-331X},
    url = {https://doi.org/10.1145/3747588},
    doi = {10.1145/3747588},
    journal = {ACM Trans. Softw. Eng. Methodol.},
    month = jan,
    articleno = {58},
    numpages = {72}
    }

@misc{deepseekai2025deepseekv3technicalreport,
      title={DeepSeek-V3 Technical Report}, 
      author={DeepSeek-AI and Aixin Liu and Bei Feng and Bing Xue and Bingxuan Wang and Bochao Wu and Chengda Lu and Chenggang Zhao and Chengqi Deng and Chenyu Zhang and Chong Ruan and Damai Dai and Daya Guo and Dejian Yang and Deli Chen and Dongjie Ji and Erhang Li and Fangyun Lin and Fucong Dai and Fuli Luo and Guangbo Hao and Guanting Chen and Guowei Li and H. Zhang and Han Bao and Hanwei Xu and Haocheng Wang and Haowei Zhang and Honghui Ding and Huajian Xin and Huazuo Gao and Hui Li and Hui Qu and J. L. Cai and Jian Liang and Jianzhong Guo and Jiaqi Ni and Jiashi Li and Jiawei Wang and Jin Chen and Jingchang Chen and Jingyang Yuan and Junjie Qiu and Junlong Li and Junxiao Song and Kai Dong and Kai Hu and Kaige Gao and Kang Guan and Kexin Huang and Kuai Yu and Lean Wang and Lecong Zhang and Lei Xu and Leyi Xia and Liang Zhao and Litong Wang and Liyue Zhang and Meng Li and Miaojun Wang and Mingchuan Zhang and Minghua Zhang and Minghui Tang and Mingming Li and Ning Tian and Panpan Huang and Peiyi Wang and Peng Zhang and Qiancheng Wang and Qihao Zhu and Qinyu Chen and Qiushi Du and R. J. Chen and R. L. Jin and Ruiqi Ge and Ruisong Zhang and Ruizhe Pan and Runji Wang and Runxin Xu and Ruoyu Zhang and Ruyi Chen and S. S. Li and Shanghao Lu and Shangyan Zhou and Shanhuang Chen and Shaoqing Wu and Shengfeng Ye and Shengfeng Ye and Shirong Ma and Shiyu Wang and Shuang Zhou and Shuiping Yu and Shunfeng Zhou and Shuting Pan and T. Wang and Tao Yun and Tian Pei and Tianyu Sun and W. L. Xiao and Wangding Zeng and Wanjia Zhao and Wei An and Wen Liu and Wenfeng Liang and Wenjun Gao and Wenqin Yu and Wentao Zhang and X. Q. Li and Xiangyue Jin and Xianzu Wang and Xiao Bi and Xiaodong Liu and Xiaohan Wang and Xiaojin Shen and Xiaokang Chen and Xiaokang Zhang and Xiaosha Chen and Xiaotao Nie and Xiaowen Sun and Xiaoxiang Wang and Xin Cheng and Xin Liu and Xin Xie and Xingchao Liu and Xingkai Yu and Xinnan Song and Xinxia Shan and Xinyi Zhou and Xinyu Yang and Xinyuan Li and Xuecheng Su and Xuheng Lin and Y. K. Li and Y. Q. Wang and Y. X. Wei and Y. X. Zhu and Yang Zhang and Yanhong Xu and Yanhong Xu and Yanping Huang and Yao Li and Yao Zhao and Yaofeng Sun and Yaohui Li and Yaohui Wang and Yi Yu and Yi Zheng and Yichao Zhang and Yifan Shi and Yiliang Xiong and Ying He and Ying Tang and Yishi Piao and Yisong Wang and Yixuan Tan and Yiyang Ma and Yiyuan Liu and Yongqiang Guo and Yu Wu and Yuan Ou and Yuchen Zhu and Yuduan Wang and Yue Gong and Yuheng Zou and Yujia He and Yukun Zha and Yunfan Xiong and Yunxian Ma and Yuting Yan and Yuxiang Luo and Yuxiang You and Yuxuan Liu and Yuyang Zhou and Z. F. Wu and Z. Z. Ren and Zehui Ren and Zhangli Sha and Zhe Fu and Zhean Xu and Zhen Huang and Zhen Zhang and Zhenda Xie and Zhengyan Zhang and Zhewen Hao and Zhibin Gou and Zhicheng Ma and Zhigang Yan and Zhihong Shao and Zhipeng Xu and Zhiyu Wu and Zhongyu Zhang and Zhuoshu Li and Zihui Gu and Zijia Zhu and Zijun Liu and Zilin Li and Ziwei Xie and Ziyang Song and Ziyi Gao and Zizheng Pan},
      year={2025},
      eprint={2412.19437},
      archivePrefix={arXiv},
      primaryClass={cs.CL},
      url={https://arxiv.org/abs/2412.19437}, 
}

@article{Guo_2025,
   title={DeepSeek-R1 incentivizes reasoning in LLMs through reinforcement learning},
   volume={645},
   ISSN={1476-4687},
   url={http://dx.doi.org/10.1038/s41586-025-09422-z},
   DOI={10.1038/s41586-025-09422-z},
   number={8081},
   journal={Nature},
   publisher={Springer Science and Business Media LLC},
   author={Guo, Daya and Yang, Dejian and Zhang, Haowei and Song, Junxiao and Wang, Peiyi and Zhu, Qihao and Xu, Runxin and Zhang, Ruoyu and Ma, Shirong and Bi, Xiao and Zhang, Xiaokang and Yu, Xingkai and Wu, Yu and Wu, Z. F. and Gou, Zhibin and Shao, Zhihong and Li, Zhuoshu and Gao, Ziyi and Liu, Aixin and Xue, Bing and Wang, Bingxuan and Wu, Bochao and Feng, Bei and Lu, Chengda and Zhao, Chenggang and Deng, Chengqi and Ruan, Chong and Dai, Damai and Chen, Deli and Ji, Dongjie and Li, Erhang and Lin, Fangyun and Dai, Fucong and Luo, Fuli and Hao, Guangbo and Chen, Guanting and Li, Guowei and Zhang, H. and Xu, Hanwei and Ding, Honghui and Gao, Huazuo and Qu, Hui and Li, Hui and Guo, Jianzhong and Li, Jiashi and Chen, Jingchang and Yuan, Jingyang and Tu, Jinhao and Qiu, Junjie and Li, Junlong and Cai, J. L. and Ni, Jiaqi and Liang, Jian and Chen, Jin and Dong, Kai and Hu, Kai and You, Kaichao and Gao, Kaige and Guan, Kang and Huang, Kexin and Yu, Kuai and Wang, Lean and Zhang, Lecong and Zhao, Liang and Wang, Litong and Zhang, Liyue and Xu, Lei and Xia, Leyi and Zhang, Mingchuan and Zhang, Minghua and Tang, Minghui and Zhou, Mingxu and Li, Meng and Wang, Miaojun and Li, Mingming and Tian, Ning and Huang, Panpan and Zhang, Peng and Wang, Qiancheng and Chen, Qinyu and Du, Qiushi and Ge, Ruiqi and Zhang, Ruisong and Pan, Ruizhe and Wang, Runji and Chen, R. J. and Jin, R. L. and Chen, Ruyi and Lu, Shanghao and Zhou, Shangyan and Chen, Shanhuang and Ye, Shengfeng and Wang, Shiyu and Yu, Shuiping and Zhou, Shunfeng and Pan, Shuting and Li, S. S. and Zhou, Shuang and Wu, Shaoqing and Yun, Tao and Pei, Tian and Sun, Tianyu and Wang, T. and Zeng, Wangding and Liu, Wen and Liang, Wenfeng and Gao, Wenjun and Yu, Wenqin and Zhang, Wentao and Xiao, W. L. and An, Wei and Liu, Xiaodong and Wang, Xiaohan and Chen, Xiaokang and Nie, Xiaotao and Cheng, Xin and Liu, Xin and Xie, Xin and Liu, Xingchao and Yang, Xinyu and Li, Xinyuan and Su, Xuecheng and Lin, Xuheng and Li, X. Q. and Jin, Xiangyue and Shen, Xiaojin and Chen, Xiaosha and Sun, Xiaowen and Wang, Xiaoxiang and Song, Xinnan and Zhou, Xinyi and Wang, Xianzu and Shan, Xinxia and Li, Y. K. and Wang, Y. Q. and Wei, Y. X. and Zhang, Yang and Xu, Yanhong and Li, Yao and Zhao, Yao and Sun, Yaofeng and Wang, Yaohui and Yu, Yi and Zhang, Yichao and Shi, Yifan and Xiong, Yiliang and He, Ying and Piao, Yishi and Wang, Yisong and Tan, Yixuan and Ma, Yiyang and Liu, Yiyuan and Guo, Yongqiang and Ou, Yuan and Wang, Yuduan and Gong, Yue and Zou, Yuheng and He, Yujia and Xiong, Yunfan and Luo, Yuxiang and You, Yuxiang and Liu, Yuxuan and Zhou, Yuyang and Zhu, Y. X. and Huang, Yanping and Li, Yaohui and Zheng, Yi and Zhu, Yuchen and Ma, Yunxian and Tang, Ying and Zha, Yukun and Yan, Yuting and Ren, Z. Z. and Ren, Zehui and Sha, Zhangli and Fu, Zhe and Xu, Zhean and Xie, Zhenda and Zhang, Zhengyan and Hao, Zhewen and Ma, Zhicheng and Yan, Zhigang and Wu, Zhiyu and Gu, Zihui and Zhu, Zijia and Liu, Zijun and Li, Zilin and Xie, Ziwei and Song, Ziyang and Pan, Zizheng and Huang, Zhen and Xu, Zhipeng and Zhang, Zhongyu and Zhang, Zhen},
   year={2025},
   month=Sept, pages={633–638} }

@misc{kimiteam2026kimik3openfrontier,
      title={Kimi K3: Open Frontier Intelligence}, 
      author={{Kimi Team} and Tongtong Bai and Yifan Bai and Yiping Bao and M. C. and Jianfeng Cai and Xinyuan Cai and Peizhou Cao and Yuxuan Cao and Ziwei Chai and Y. Charles and H. S. Che and Guanduo Chen and Guangyu Chen and Guanzheng Chen and Huarong Chen and Jia Chen and Jianlong Chen and Jun Chen and Kexin Chen and Peng Chen and Ruijue Chen and Wentao Chen and Xin Chen and Yang Chen and Yanru Chen and Yifei Chen and Yingjiang Chen and Yuankun Chen and Yujie Chen and Yutian Chen and Zhirong Chen and Dazhi Cheng and Yean Cheng and Jialei Cui and Jingbing Cui and Anqi Dai and Jiaqi Deng and Hao Ding and Rui Ding and Shaofeng Ding and Mengfan Dong and Mengnan Dong and Yuhao Dong and Yuxin Dong and Angang Du and Chenzhuang Du and Dikang Du and Jusen Du and Yulun Du and Yu Fan and Jing Feng and Qiulin Feng and Yichen Feng and Kelin Fu and Qiang Fu and Fuxuan Gao and Hongcheng Gao and Jingyue Gao and Tong Gao and Weijia Gao and Shangyi Geng and Jie Gong and Linhu Gong and Shengao Gong and Xiaochen Gong and Qizheng Gu and Yicheng Gu and Shuhao Guan and Haiqing Guo and Shiqi Guo and Xiang Guo and Zhengyan Guo and Beixi Hao and Wenxin Hao and Xiaoru Hao and Dailan He and Haotian He and Lehan He and Qi He and Weiran He and Xinran He and Xinyi He and Yibo He and Yunjia He and Chao Hong and Tiange Hong and Hao Hu and Jiaxi Hu and Ruikun Hu and Weiming Hu and Yangyang Hu and Zhenxing Hu and Liang Hua and Jinbin Huang and Ke Huang and Ruiyuan Huang and Siying Huang and Weixiao Huang and Yan Huang and Zhengjie Huang and Zhiqi Huang and Yulong Hui and Chaobo Jia and Yutong Jiang and Zhejun Jiang and Zuoyou Jiang and Wenyi Jin and Xinyi Jin and Yu Jing and Huanjun Kong and Guokun Lai and Aidi Li and Cheng Li and Chengyuan Li and Cong Li and Fang Li and Guanyu Li and Haoyang Li and Jia Li and Junxiong Li and Lei Li and Letian Li and Lincan Li and Weihong Li and Wentao Li and Xintong Li and Yang Li and Yishen Li and Yiwei Li and Yuxiao Li and Zhaowei Li and Zhaoxi Li and Zheming Li and Zhengxiao Li and Zhiyuan Li and Jiawei Lin and Xiaohan Lin and Yibo Lin and Zichao Lin and Ziyan Lin and Bill Liu and Boxiao Liu and Chuan Liu and Liang Liu and Shaowei Liu and Shudong Liu and Shuran Liu and Tianwei Liu and Weizhou Liu and Yangyang Liu and Yanming Liu and Yibo Liu and Yipeng Liu and Zhengying Liu and Zhiheng Liu and Enzhe Lu and Haoyu Lu and Linqiang Lu and Tingzhan Lu and Zhiyuan Lu and Aotian Luo and G. Luo and Junyu Luo and Yifan Luo and B. Lyu and Wenzhou Lyu and Shaoguang Mao and Yuan Mei and Xin Men and Minqing Ni and Yixuan Niu and Siyuan Pan and Shujun Peng and Zhangyang Qi and Ruoyu Qin and ZeChao Qin and Zeyu Qin and Haiquan Qiu and Jianxin Qiu and Jiezhong Qiu and Bowen Qu and Yuhao Qu and Zeyu Shang and Youbo Shao and Han Shen and Jincheng Shi and Juanfeng Shi and Lidong Shi and Shengyuan Shi and Wingchun Siu and Pengwei Song and Xiaoxi Song and Jianlin Su and Yunfeng Su and Zhaochen Su and Lin Sui and Jingsong Sun and Junyao Sun and Shaoning Sun and Shuzhe Sun and Tongyu Sun and Yujun Sun and Yunpeng Tai and Chuning Tang and Heyi Tang and Sirui Tang and Zecheng Tang and Chaoran Tian and Rongpeng Tian and Yu Tian and Wei Tu and Chensi Wang and Chuang Wang and Chunjie Wang and Dinglu Wang and Feng Wang and Hailong Wang and Haiming Wang and Hao Wang and Hao Wang and Huaqing Wang and Hui Wang and Jiayi Wang and Jinglong Wang and Jinhong Wang and Jiuzheng Wang and Linian Wang and Shaobo Wang and Shenzhi Wang and Shuyi Wang and Si Wang and Siyuan Wang and Tianfu Wang and Wenjue Wang and Xingran Wang and Xinmei Wang and Xinyuan Wang and Xusheng Wang and Yalin Wang and Yangkun Wang and Yao Wang and Yaoyu Wang and Yejie Wang and Yiqin Wang and Yucheng Wang and Yuzhi Wang and Zhaoji Wang and Zhaowei Wang and Zhengtao Wang and Zhenhao Wang and Zhongsheng Wang and Zifan Wang and Chu Wei and Ming Wei and Shouxin Wei and Zichen Wen and Fan Wu and Haoning Wu and Rucong Wu and Wenhao Wu and Xiaoxue Wu and Yingcong Wu and Yongqi Wu and Yuxin Wu and Zijian Wu and Xinglang Xian and Chenxuan Xiang and Yuye Xiang and Bocheng Xiao and Chenjun Xiao and Xin Xiao and Jin Xie and Xiaotong Xie and Yifeng Xie and Zhe Xie and Bowei Xing and Yiming Xiong and Baosheng Xu and Boyu Xu and Jiale Xu and Jianfan Xu and Jing Xu and Jinjing Xu and L. H. Xu and Qingtao Xu and Shuyao Xu and Suting Xu and Tiantian Xu and Tianxiang Xu and Weixin Xu and Xinran Xu and Yangchuan Xu and Ye Xu and Yueni Xu and Ziyao Xu and Haonan Xue and Junjie Yan and Yaoyao Yan and Fan Yang and Guangyao Yang and Hao Yang and Junwei Yang and Ruoyu Yang and Wenjie Yang and Xiaofei Yang and Xinyu Yang and Yi Yang and Yiling Yang and Ying Yang and Yuchen Yang and Zhen Yang and Zhilin Yang and Zian Yang and Zuhao Yang and Haotian Yao and Dan Ye and Haoran Ye and Wenjie Ye and Zhanbo Ye and Bohong Yin and Haoxiang Yin and Xietong Yin and Chengzhen Yu and Haozhen Yu and Longhui Yu and Shengnan Yu and Shuying Yu and Tianxiang Yu and Enming Yuan and Mengjie Yuan and Tongtian Yue and Wei Yue and Yang Yue and Dunyuan Zha and Haobing Zhan and B. H. Zhang and Dehao Zhang and Fei Zhang and Hao Zhang and Haoyuan Zhang and Huanyu Zhang and Jiapei Zhang and Jiaxuan Zhang and Jin Zhang and Kaiyi Zhang and Miaozhen Zhang and Puqi Zhang and Qinglei Zhang and Rong Zhang and Rui Zhang and Shaoshuai Zhang and Shiyi Zhang and Xiaobin Zhang and Xiaoyun Zhang and Y. Zhang and Yangkun Zhang and Ye Zhang and Yichi Zhang and Yikun Zhang and Yizhi Zhang and Yongting Zhang and Yu Zhang and Yutao Zhang and Yutong Zhang and Zheng Zhang and Zijing Zhang and Bin Zhao and Chenguang Zhao and Feifan Zhao and Jinglun Zhao and Jinxiang Zhao and Shuai Zhao and Wenshuo Zhao and Xiangyu Zhao and Xuanle Zhao and Yikai Zhao and Zijia Zhao and Haozhi Zheng and Huabin Zheng and Ruihan Zheng and Shaojie Zheng and Tengyang Zheng and Haofeng Zhong and Lei Zhong and Longguang Zhong and M. Zhou and Qiankang Zhou and Runjie Zhou and Ruozhang Zhou and Xinyu Zhou and Yiqiao Zhou and Zaida Zhou and Jinguo Zhu and Liya Zhu and Xinhao Zhu and Yangjunfeng Zhu and Yuxuan Zhu and Zhen Zhu and Chen Zhuang and Weiyu Zhuang and Xinxing Zu},
      year={2026},
      eprint={2607.24653},
      archivePrefix={arXiv},
      primaryClass={cs.CL},
      url={https://arxiv.org/abs/2607.24653}, 
}

@misc{kimiteam2026kimik2openagentic,
      title={Kimi K2: Open Agentic Intelligence}, 
      author={{Kimi Team} and Yifan Bai and Yiping Bao and Y. Charles and Cheng Chen and Guanduo Chen and Haiting Chen and Huarong Chen and Jiahao Chen and Ningxin Chen and Ruijue Chen and Yanru Chen and Yuankun Chen and Yutian Chen and Zhuofu Chen and Jialei Cui and Hao Ding and Mengnan Dong and Angang Du and Chenzhuang Du and Dikang Du and Yulun Du and Yu Fan and Yichen Feng and Kelin Fu and Bofei Gao and Chenxiao Gao and Hongcheng Gao and Peizhong Gao and Tong Gao and Yuyao Ge and Shangyi Geng and Qizheng Gu and Xinran Gu and Longyu Guan and Haiqing Guo and Jianhang Guo and Xiaoru Hao and Tianhong He and Weiran He and Wenyang He and Yunjia He and Chao Hong and Hao Hu and Yangyang Hu and Zhenxing Hu and Weixiao Huang and Zhiqi Huang and Zihao Huang and Tao Jiang and Zhejun Jiang and Xinyi Jin and Yongsheng Kang and Guokun Lai and Cheng Li and Fang Li and Haoyang Li and Ming Li and Wentao Li and Yang Li and Yanhao Li and Yiwei Li and Zhaowei Li and Zheming Li and Hongzhan Lin and Xiaohan Lin and Zongyu Lin and Chengyin Liu and Chenyu Liu and Hongzhang Liu and Jingyuan Liu and Junqi Liu and Liang Liu and Shaowei Liu and T. Y. Liu and Tianwei Liu and Weizhou Liu and Yangyang Liu and Yibo Liu and Yiping Liu and Yue Liu and Zhengying Liu and Enzhe Lu and Haoyu Lu and Lijun Lu and Yashuo Luo and Shengling Ma and Xinyu Ma and Yingwei Ma and Shaoguang Mao and Jie Mei and Xin Men and Yibo Miao and Siyuan Pan and Yebo Peng and Ruoyu Qin and Zeyu Qin and Bowen Qu and Zeyu Shang and Lidong Shi and Shengyuan Shi and Feifan Song and Jianlin Su and Zhengyuan Su and Lin Sui and Xinjie Sun and Flood Sung and Yunpeng Tai and Heyi Tang and Jiawen Tao and Qifeng Teng and Chaoran Tian and Chensi Wang and Dinglu Wang and Feng Wang and Hailong Wang and Haiming Wang and Jianzhou Wang and Jiaxing Wang and Jinhong Wang and Shengjie Wang and Shuyi Wang and Si Wang and Xinyuan Wang and Yao Wang and Yejie Wang and Yiqin Wang and Yuxin Wang and Yuzhi Wang and Zhaoji Wang and Zhengtao Wang and Zhengtao Wang and Zhexu Wang and Chu Wei and Qianqian Wei and Haoning Wu and Wenhao Wu and Xingzhe Wu and Yuxin Wu and Chenjun Xiao and Jin Xie and Xiaotong Xie and Weimin Xiong and Boyu Xu and Jinjing Xu and L. H. Xu and Lin Xu and Suting Xu and Weixin Xu and Xinran Xu and Yangchuan Xu and Ziyao Xu and Jing Xu and Jing Xu and Junjie Yan and Yuzi Yan and Hao Yang and Xiaofei Yang and Yi Yang and Ying Yang and Zhen Yang and Zhilin Yang and Zonghan Yang and Haotian Yao and Xingcheng Yao and Wenjie Ye and Zhuorui Ye and Bohong Yin and Longhui Yu and Enming Yuan and Hongbang Yuan and Mengjie Yuan and Siyu Yuan and Haobing Zhan and Dehao Zhang and Hao Zhang and Wanlu Zhang and Xiaobin Zhang and Yadong Zhang and Yangkun Zhang and Yichi Zhang and Yizhi Zhang and Yongting Zhang and Yu Zhang and Yutao Zhang and Yutong Zhang and Zheng Zhang and Haotian Zhao and Yikai Zhao and Zijia Zhao and Huabin Zheng and Shaojie Zheng and Longguang Zhong and Jianren Zhou and Xinyu Zhou and Zaida Zhou and Jinguo Zhu and Zhen Zhu and Weiyu Zhuang and Xinxing Zu},
      year={2026},
      eprint={2507.20534},
      archivePrefix={arXiv},
      primaryClass={cs.LG},
      url={https://arxiv.org/abs/2507.20534}, 
}

@misc{glm5team2026glm5vibecodingagentic,
      title={GLM-5: from Vibe Coding to Agentic Engineering}, 
      author={GLM-5-Team and Aohan Zeng and Xin Lv and Zhenyu Hou and Zhengxiao Du and Qinkai Zheng and Bin Chen and Da Yin and Chendi Ge and Chenghua Huang and Chengxing Xie and Chenzheng Zhu and Congfeng Yin and Cunxiang Wang and Gengzheng Pan and Hao Zeng and Haoke Zhang and Haoran Wang and Huilong Chen and Jiajie Zhang and Jian Jiao and Jiaqi Guo and Jingsen Wang and Jingzhao Du and Jinzhu Wu and Kedong Wang and Lei Li and Lin Fan and Lucen Zhong and Mingdao Liu and Mingming Zhao and Pengfan Du and Qian Dong and Rui Lu and Shuang-Li and Shulin Cao and Song Liu and Ting Jiang and Xiaodong Chen and Xiaohan Zhang and Xuancheng Huang and Xuezhen Dong and Yabo Xu and Yao Wei and Yifan An and Yilin Niu and Yitong Zhu and Yuanhao Wen and Yukuo Cen and Yushi Bai and Zhongpei Qiao and Zihan Wang and Zikang Wang and Zilin Zhu and Ziqiang Liu and Zixuan Li and Bojie Wang and Bosi Wen and Can Huang and Changpeng Cai and Chao Yu and Chen Li and Chengwei Hu and Chenhui Zhang and Dan Zhang and Daoyan Lin and Dayong Yang and Di Wang and Ding Ai and Erle Zhu and Fangzhou Yi and Feiyu Chen and Guohong Wen and Hailong Sun and Haisha Zhao and Haiyi Hu and Hanchen Zhang and Hanrui Liu and Hanyu Zhang and Hao Peng and Hao Tai and Haobo Zhang and He Liu and Hongwei Wang and Hongxi Yan and Hongyu Ge and Huan Liu and Huanpeng Chu and Jia'ni Zhao and Jiachen Wang and Jiajing Zhao and Jiamin Ren and Jiapeng Wang and Jiaxin Zhang and Jiayi Gui and Jiayue Zhao and Jijie Li and Jing An and Jing Li and Jingwei Yuan and Jinhua Du and Jinxin Liu and Junkai Zhi and Junwen Duan and Kaiyue Zhou and Kangjian Wei and Ke Wang and Keyun Luo and Laiqiang Zhang and Leigang Sha and Liang Xu and Lindong Wu and Lintao Ding and Lu Chen and Minghao Li and Nianyi Lin and Pan Ta and Qiang Zou and Rongjun Song and Ruiqi Yang and Shangqing Tu and Shangtong Yang and Shaoxiang Wu and Shengyan Zhang and Shijie Li and Shuang Li and Shuyi Fan and Wei Qin and Wei Tian and Weining Zhang and Wenbo Yu and Wenjie Liang and Xiang Kuang and Xiangmeng Cheng and Xiangyang Li and Xiaoquan Yan and Xiaowei Hu and Xiaoying Ling and Xing Fan and Xingye Xia and Xinyuan Zhang and Xinze Zhang and Xirui Pan and Xu Zou and Xunkai Zhang and Yadi Liu and Yandong Wu and Yanfu Li and Yidong Wang and Yifan Zhu and Yijun Tan and Yilin Zhou and Yiming Pan and Ying Zhang and Yinpei Su and Yipeng Geng and Yong Yan and Yonglin Tan and Yuean Bi and Yuhan Shen and Yuhao Yang and Yujiang Li and Yunan Liu and Yunqing Wang and Yuntao Li and Yurong Wu and Yutao Zhang and Yuxi Duan and Yuxuan Zhang and Zezhen Liu and Zhengtao Jiang and Zhenhe Yan and Zheyu Zhang and Zhixiang Wei and Zhuo Chen and Zhuoer Feng and Zijun Yao and Ziwei Chai and Ziyuan Wang and Zuzhou Zhang and Bin Xu and Minlie Huang and Hongning Wang and Juanzi Li and Yuxiao Dong and Jie Tang},
      year={2026},
      eprint={2602.15763},
      archivePrefix={arXiv},
      primaryClass={cs.LG},
      url={https://arxiv.org/abs/2602.15763}, 
}

@misc{deepseekai2026deepseekv4highlyefficientmilliontoken,
      title={DeepSeek-V4: Towards Highly Efficient Million-Token Context Intelligence}, 
      author={DeepSeek-AI and Anyi Xu and Bangcai Lin and Bing Xue and Bingxuan Wang and Bingzheng Xu and Bochao Wu and Bowei Zhang and Chaofan Lin and Chen Dong and Chenchen Ling and Chengda Lu and Chenggang Zhao and Chengqi Deng and Chengyu Hou and Chenhao Xu and Chenze Shao and Chong Ruan and Conner Sun and Damai Dai and Daya Guo and Dejian Yang and Deli Chen and Donghao Li and Dongjie Ji and Erhang Li and Fang Wei and Fangyun Lin and Fangzhou Yuan and Feiyu Xia and Fucong Dai and Guangbo Hao and Guanting Chen and Guoai Cao and Guolai Meng and Guowei Li and Han Yu and Han Zhang and Hanwei Xu and Hao Li and Haofen Liang and Haoling Zhang and Haoming Luo and Haoran Wei and Haotian Yuan and Haowei Zhang and Haowen Luo and Haoyu Chen and Haozhe Ji and Hengqing Zhang and Honghui Ding and Hongxuan Tang and Huanqi Cao and Huazuo Gao and Hui Qu and Hui Zeng and J Yang and JQ Zhu and Jia Luo and Jia Song and Jia Yu and Jialiang Huang and Jialu Cai and Jian Liang and Jiangting Zhou and Jiasheng Ye and Jiashi Li and Jiaxin Xu and Jiewen Hu and Jieyu Yang and Jin Chen and Jin Yan and Jingchang Chen and Jingli Zhou and Jingting Xiang and Jingyang Yuan and Jingyuan Cheng and Jingzi Zhou and Jinhua Zhu and Jiping Yu and Joseph Sun and Jun Ran and Junguang Jiang and Junjie Qiu and Junlong Li and Junmin Zheng and Junxiao Song and Kai Dong and Kaige Gao and Kang Guan and Kexing Zhou and Kezhao Huang and Kuai Yu and Lean Wang and Lecong Zhang and Lei Wang and Leyi Xia and Li Zhang and Liang Zhao and Lihua Guo and Lingxiao Luo and Linwang Ma and Linyan Zhu and Litong Wang and Liyu Cai and Liyue Zhang and Longhao Chen and MS Di and MY Xu and Max Mei and Miaojun Wang and Mingchuan Zhang and Minghua Zhang and Minghui Tang and Mingming Li and Mingxu Zhou and Minmin Han and Ning Wang and Panpan Huang and Panpan Wang and Peixin Cong and Peiyi Wang and Peng Zhang and Qiancheng Wang and Qihao Zhu and Qingyang Li and Qinyu Chen and Qiushi Du and Qiwei Jiang and Rui Tian and Ruifan Xu and Ruijie Lu and Ruiling Xu and Ruiqi Ge and Ruisong Zhang and Ruizhe Pan and Runji Wang and Runqian Chen and Runqiu Yin and Runxin Xu and Ruomeng Shen and Ruoyu Zhang and Ruyi Chen and SH Liu and Shanghao Lu and Shangmian Sun and Shangyan Zhou and Shanhuang Chen and Shaofei Cai and Shaoheng Nie and Shaoqing Wu and Shaoyuan Chen and Shengding Hu and Shengyu Liu and Shiqiang Hu and Shirong Ma and Shiyu Wang and Shuiping Yu and Shunfeng Zhou and Shuting Pan and Shuying Yu and Songyang Zhou and Tao Ni and Tao Yun and Tian Jin and Tian Pei and Tian Ye and Tianle Lin and Tianran Ji and Tianyi Cui and Tianyuan Yue and Tingting Yu and Tun Wang and W Zhang and WL Xiao and Wangding Zeng and Wei An and Weilin Zhao and Wen Liu and Wenfeng Liang and Wenjie Pang and Wenjing Luo and Wenjing Yao and Wenjun Gao and Wenkai Yang and Wenlve Huang and Wenqing Hou and Wentao Zhang and Wenting Ma and Xi Gao and Xiang He and Xiangwen Wang and Xianzu Wang and Xiao Bi and Xiaodong Liu and Xiaohan Wang and Xiaokang Chen and Xiaokang Zhang and Xiaotao Nie and Xiaowen Sun and Xiaoxiang Wang and Xin Cheng and Xin Liu and Xin Xie and Xingchao Liu and Xingchen Liu and Xingkai Yu and Xingyou Li and Xinyu Yang and Xinyu Zhang and Xu Chen and Xuanyu Wang and Xuecheng Su and Xueyin Chen and Xuheng Lin and Xuwei Fu and YC Yan and YQ Wang and YW Ma and Yanfeng Luo and Yang Zhang and Yanhong Xu and Yanru Ma and Yanwen Huang and Yao Li and Yao Li and Yao Xu and Yao Zhao and Yaofeng Sun and Yaohui Wang and Yi Qian and Yi Shao and Yi Yu and Yichao Zhang and Yifan Ding and Yifan Shi and Yijia Wu and Yiliang Xiong and Yiling Ma and Ying He and Ying Tang and Ying Zhou and Yingjia Luo and Yinmin Zhong and Yishi Piao and Yisong Wang and Yixiang Zhang and Yixiao Chen and Yixuan Tan and Yixuan Wei and Yiyang Ma and Yiyuan Liu and Yonglun Yang and Yongqiang Guo and Yongtong Wu and Yu Wu and YuKun Li and Yuan Cheng and Yuan Ou and Yuanfan Xu and Yuanhao Li and Yuduan Wang and Yuehan Yang and Yuer Xu and Yuhan Wu and Yuhao Meng and Yuheng Zou and Yukun Zha and Yunfan Xiong and Yupeng Chen and Yuping Lin and Yuqian Cao and Yuqian Wang and Yushun Zhang and Yuting Yan and Yutong Lin and Yuxian Gu and Yuxiang Luo and Yuxiang You and Yuxuan Liu and Yuxuan Zhou and Yuyang Zhou and Yuzhen Huang and ZF Wu and Zehao Wang and Zehua Zhao and Zehui Ren and Zekai Zhang and Zhangli Sha and Zhe Fu and Zhe Ju and Zhean Xu and Zhenda Xie and Zhengyan Zhang and Zheren Gao and Zhewen Hao and Zhibin Gou and Zhicheng Ma and Zhigang Yan and Zhihong Shao and Zhixian Huang and Zhixuan Chen and Zhiyu Wu and Zhizhou Ren and Zhongyu Wu and Zhuoshu Li and Zhuping Zhang and Zian Xu and Zihao Wang and Zihua Qu and Zihui Gu and Zijia Zhu and Zilin Li and Zipeng Zhang and Ziwei Xie and Ziyi Gao and Ziyi Wan and Zizheng Pan and Zongqing Yao},
      year={2026},
      eprint={2606.19348},
      archivePrefix={arXiv},
      primaryClass={cs.CL},
      url={https://arxiv.org/abs/2606.19348}, 
}

@misc{qwen38,
    title = {Qwen3.8-Max: A New Bar for Coding and Cowork},
    url = {https://qwen.ai/blog?id=qwen3.8},
    author = {{Qwen Team}},
    month = {August},
    year = {2026}
}

@misc{qwen3.6-35b-a3b,
  title = {{Qwen3.6-35B-A3B}: Agentic Coding Power, Now Open to All},
  url = {https://qwen.ai/blog?id=qwen3.6-35b-a3b},
  author = {{Qwen Team}},
  month = {April},
  year = {2026}
}

@inproceedings{evalplus,
  title = {Is Your Code Generated by Chat{GPT} Really Correct? Rigorous Evaluation of Large Language Models for Code Generation},
  author = {Liu, Jiawei and Xia, Chunqiu Steven and Wang, Yuyao and Zhang, Lingming},
  booktitle = {Thirty-seventh Conference on Neural Information Processing Systems},
  year = {2023},
  url = {https://openreview.net/forum?id=1qvx610Cu7},
}

@inproceedings{yang2024sweagent,
  title={{SWE}-agent: Agent-Computer Interfaces Enable Automated Software Engineering},
  author={John Yang and Carlos E Jimenez and Alexander Wettig and Kilian Lieret and Shunyu Yao and Karthik R Narasimhan and Ofir Press},
  booktitle={The Thirty-eighth Annual Conference on Neural Information Processing Systems},
  year={2024},
  url={https://arxiv.org/abs/2405.15793}
}

@misc{qiu2026designqwen38nextarchitectureevaluation,
      title={On the Design of Qwen3.8-Next Architecture: Evaluation, Efficiency, and Training Stability}, 
      author={Zihan Qiu and Zekun Wang and Xiao Li and Yanpeng Li and Yang Xu and Yixuan Wang and Huaqing Zhang and Rui Men and Bochao Mao and Chengruidong Zhang and Fan Zhou and Hao Luo and Haofeng Huang and Haoran Lian and Haoyan Huang and Hongqing Chen and Jianwei Zhang and Jing Xu and Junjie Wang and Langshi Chen and Liangyu Wang and Linlang Jiang and Man Yuan and Minmin Sun and Peng Jin and Siqi Zhang and Siyu Wang and Xingzhang Ren and Yakai Wang and Yi Zhang and Yiming Dong and Yizhong Cao and Yubo Ma and Yunfei Mao and Bo Zheng and Dayiheng Liu},
      year={2026},
      eprint={2608.30320},
      archivePrefix={arXiv},
      primaryClass={cs.CL},
      url={https://arxiv.org/abs/2608.30320}, 
}

@misc{deepseekai2026deepseekv41flashpushinglimitskv,
      title={DeepSeek-V4.1-Flash: Pushing the Limits of KV Cache Compression}, 
      author={DeepSeek-AI and Anyi Xu and B. Li and Bangcai Lin and Bing Xue and BingCheng Xian and Bingzheng Xu and Bochao Wu and Bowei Zhang and Boyi Deng and C. C. Yu and Chao Jin and Chaofan Lin and Chen Dong and Chenbing Wang and Chenfan Feng and Chengda Lu and Chenggang Zhao and Chengqi Deng and Chengyuan Zhang and Chenhao Xu and Chenqi Zhao and Chenze Shao and Chuhao Wang and Chuqi Zhang and Damai Dai and Dejian Yang and Deli Chen and Di Huang and Di Wu and Donghao Li and Erhang Li and Eric Fu and F. Zhou and Fangwei Zhou and Fangyun Lin and Fangzhou Yuan and Feiyu Xia and Fucong Dai and Guangbo Hao and Guanglin Li and Guanting Chen and Guoai Cao and Guofan Fan and Guolai Meng and Guowei Li and Haichuan Zhang and Haiyang Ma and Haiyang Shen and Han Li and Han Yu and Han Zhang and Hangyuan Deng and Hanwei Xu and Hanxiang Xu and Hanxun Zhong and Hao Guo and Hao Jiang and Hao Li and Hao Qin and Haodong Wen and Haofen Liang and Haofeng Huang and Haohua Liu and Haoling Zhang and Haoming Luo and Haoran Yang and Haotian Xu and Haotian Yuan and Haoting Huang and Haowen Luo and Haoyang Cai and Haoyu Chen and Haozhe Ji and Hengran Zhang and Hengrui Wang and Hengxu Wu and Honghui Ding and Hongxuan Tang and Huadong Wang and Huanqi Cao and Huazuo Gao and Hui Qu and Hui Zeng and J. Yang and J. H. Jin and J. H. Zhang and J. X. Zou and Jia Yu and Jiahui Zhou and Jiajun Chen and Jialiang Huang and Jialin Zhao and Jiamin Tang and Jian Zhou and Jianan Tong and Jianwen Li and Jiaqi Zhu and Jiarui Wang and Jiasheng Ye and Jiashi Li and Jiaxin Xu and Jiaying Ding and Jibai Lu and Jiewen Hu and Jin Yan and Jincheng Zhai and Jingchang Chen and Jingcheng Hu and Jingli Zhou and Jingsheng Xu and Jingting Xiang and Jingyan Yun and Jingyang Yuan and Jingyuan Cheng and Jinhua Zhu and Jinpeng Wang and Jinyi Chen and Jinyi Hu and Jiping Yu and Jueliang Guo and Junbo Pei and Junbo Sun and Junguang Jiang and Junjie Qiu and Junkang Zhou and Junqi Liu and Junren Li and Junxian Li and Junxiao Song and Junyi Guo and Kai Dong and Kaifeng Chen and Kaige Gao and Kang Guan and Kangdong Yuan and Ke Hong and Ke Xu and Kefan Zhao and Kexin Ji and Kexin Zhang and Kexing Zhou and Kuai Yu and Lan Zhang and Lean Wang and Lecong Zhang and Lei Wang and Letian Gao and Liang Zhao and Liansheng Xu and Lihua Guo and Lingxiao Luo and Lingyue Fu and Litao Deng and Litong Wang and Liyue Zhang and Longhao Chen and Lu Chen and Luotian Huang and Luyao Ma and Luyao Wang and M. S. Di and Max Mei and Menghao Ye and Miao Cui and Mingchuan Zhang and Minghua Zhang and Minghui Tang and Mingjing Zhang and Mingqi Wei and Mingshu Chen and Mingxing Liu and Mingxu Zhou and Mingyu Xu and Mingyu Yang and Mingze Wang and Muyang Chen and Ni Shentu and Ning Wang and Niufang Ning and Panpan Huang and Peixin Cong and Peiyi Wang and Peiyuan Xin and Pengfei Ren and Pengfei Yan and Pengle Zhang and Qi Kang and Qi Tang and Qiancheng Wang and Qiang Li and Qihao Zhu and Qingyang Li and Qinyu Chen and Qiushi Du and Qizhou Guo and Rongxian Xu and Rui Ding and Rui Hu and Rui Tian and Rui Yu and Ruidong Zhu and Ruifan Xu and Ruihan Yang and Ruihang Xia and Ruijie Lu and Ruilin Geng and Ruipeng Hong and Ruiqi Ge and Ruisong Zhang and Ruize Sun and Ruizhe Pan and Runji Wang and Runqian Chen and Runxin Xu and Ruohong Tian and Ruomeng Shen and Ruoyu Zhang and Ryan X. and S. H. Liu and Shanghao Lu and Shangyan Zhou and Shanhuang Chen and Shaofei Cai and Shaoheng Nie and Shaoyuan Chen and Shengding Hu and Shengkai Lin and Shengwen Ran and Shengyu Liu and Shengyuan Jia and Shi Bai and Shi Feng and Shicheng Xu and Shichun Liu and Shiqiang Hu and Shirong Ma and Shiyu Wang and Shiyuan Feng and Shufan Gong and Shuhan Lin and Shuiping Yu and Shunfeng Zhou and Shuo Yang and Shuomeng Wang and Shuting Guo and Shuting Pan and Shuying Yu and Sinuo Cao and Siyi Lin and Sizhe Chen and Songyang Chen and Songyang Zhou and Tao Ni and Tao Yun and Tian Jin and Tian Pei and Tian Ye and Tianle Lin and Tianran Ji and Tianyi Cui and Tianyuan Yue and Tingting Yu and Tongrui Xiong and Wangding Zeng and Wei Liu and Wei Zhang and Weibin Xu and Weihao Zeng and Weilin Zhao and Wen Liu and Wenfeng Liang and Wenjie Pang and Wenjing Luo and Wenjing Yao and Wenjun Gao and Wenkai Shao and Wenkai Yang and Wenli Zhang and Wenlu Wang and Wenlve Huang and Wenqian Yan and Wentao Zhang and Xi Gao and Xiang He and Xiang Li and Xiangli Li and Xiangwen Wang and Xiangying Zhang and Xiankui Wei and Xiao Bi and Xiaodong Liu and Xiaohan Wang and Xiaojian Qu and Xiaokang Chen and Xiaokang Zhang and Xiaotao Nie and Xiaoyao Zou and Xiaoyuan Li and Xicheng Guo and Xieting Chu and Xin Cheng and Xin Liu and Xin Xie and Xinbo Xu and Xingchao Liu and Xingchen Liu and Xingkai Yu and Xingyou Li and Xintong Yao and Xinyang Chen and Xinyong Jiang and Xinyu Yang and Xinyu Yang and Xu Chen and Xuanyu Wang and Xubei Zhong and Xuecheng Su and Xuejie Liu and Xuheng Lin and Xujie Fan and Xuncheng Zhao and Xuwei Fu and Y. C. Yan and Y. H. Jiang and Y. T. Wu and Y. W. M. and Y. Z. Wang and Yafei Gao and Yang Yang and Yang Zhang and Yanru Ma and Yanwen Huang and Yao Li and Yao Li and Yao Meng and Yao Zhao and Yaofeng Sun and Yaohui Wang and Yaoyang Ye and Yehang Yin and Yexinrui Wu and Yi Qian and Yi Tao and Yi Yu and Yichao Zhang and Yichen Jiang and Yicheng Wang and Yifan Ding and Yifan Shi and Yifeng Peng and Yifeng Zhai and Yijia Wu and Yiliang Xiong and Yilun Wang and Ying He and Ying Zhou and Yingjia Luo and Yinmin Zhong and Yiping Wang and Yisong Wang and Yixiang Zhang and Yixiao Chen and Yixuan Tan and Yixuan Wei and Yiyang Ma and Yiyao Yang and Yiyuan Liu and Yizai Cai and Yizhen Wei and Yizhi Wang and Yonglun Yang and Yongqi Zhuo and Yongqiang Guo and Yongtong Wu and Yu Wu and Yu Zhang and Yuan Bian and Yuan Cheng and Yuan Ou and Yuan Sun and Yuanfan Xu and Yuanhang Sun and Yuanhao Li and Yuchen Liu and Yuchen Yao and Yudong Han and Yuduan Wang and Yuhan Wu and Yuhao Meng and Yuheng Zou and YuKun Li and Yunchuan Wang and Yunfan Xiao and Yunfan Xiong and Yupeng Chen and Yuqian Cao and Yuqian Wang and Yuqing Chen and Yushun Zhang and Yutong Lin and Yuwei Xiao and Yuxian Gu and Yuxiang Chen and Yuxiang Huang and Yuxiang Luo and Yuxiang You and Yuxin Chen and Yuxin Xiang and Yuxuan Liu and Yuxuan Zhou and Yuyang Zhou and Yuzhe Guo and Yuzhen Huang and Yuzhuo Bai and Z. Y. Z. and Zanlin Ni and Zehao Wang and Zehua Zhao and Zehui Ren and Zejun Zhao and Zhangli Sha and Zhanying Wang and Zhaochen Zhang and Zhaoshuai Du and Zhe Fu and Zhean Xu and Zhenda Xie and Zheng Liu and Zhengyan Zhang and Zhenhua Dong and Zhewen Hao and Zhibang Wang and Zhibin Gou and Zhicheng Ma and Zhihao Li and Zhihong Shao and Zhihuan Huang and Zhijie Li and Zhirui Lu and Zhixian Huang and Zhixuan Chen and Zhixuan Chen and Zhixuan Pan and Zhiyu Wu and Zhizhou Ren and Zhu He and Zhuoshu Li and Zhuping Zhang and Zian Xu and Zihao Wang and Zihui Gu and Zijia Zhu and Zili Zhang and Zilin Li and Zilong Hou and Zilong Lyu and Ziqiao Wang and Ziwei Xie and Ziya Zhang and Ziyi Gao and Zizheng Pan and Zonglin Li and Zongqing Yao and Zui Chen and Zuofan Wu and Chenchen Ling and Chengyu Hou and Chong Chen and D. Li and Di Qi and Dongjie Ji and Fang Wei and Fanyi Xia and Fei Xie and Feiyi Tan and Hailong Guo and Haiyan Zhai and Hui Zhou and Huihui Tan and Huijie Li and Jia Luo and Jia Song and Jialu Cai and Jian Liang and Jiangting Zhou and Jiaqi Gao and Jiayi Shao and Jie Chen and Jieyu Yang and Jin Chen and Jingde Zhang and Jingzi Zhou and Jinqian Wang and Jinyang Liu and JinZhao Sun and Junhua Ling and Junmin Zheng and Kaicheng Yang and Ke Xu and Le Su and Leyi Xia and Liangfeng Ding and Lin Zhuo and Linwang Ma and Linyan Zhu and Liyu Cai and Luqi Yao and M. K. Zhang and Meng Li and Miao Lin and Miaojun Wang and Min Zhang and Mingming Li and Mingming Wang and Mingze Yin and Minmin Han and Nan Cao and Ning Wang and Ningxin Ma and Panpan Wang and Peihan Lin and Peng Sun and Peng Zhang and Qian Ying and Qiang Xiang and Qiao Wang and Qingmiao Mao and Qiwei Jiang and Rongli Jin and Ruyi Chen and Sha Tao and Shangmian Sun and Shaoqing Wu and Shichao Zou and Si Lei and Tianyang Zhang and Tianyu Sun and Tingting Yin and W. L. Xiao and Wei An and Wei Li and Wei Wang and Weiwei Lin and Wenqing Hou and X. Lin and Xiangfei Meng and Xianzhu Huang and Xiao Peng and Xiaoqian Li and Xiaoting Zhang and Xiaowen Sun and Xiaoxiang Wang and Xiaoyu Ye and Xinrou Zhang and Xinyu Zhang and Xue Cao and Xueyin Chen and Yanan Zhou and Yanhong Xu and Yao Xia and Yao Xu and Yi Shao and Yihong Zhang and Yiling Ma and Ying Tang and Yining Lou and Yiru Chen and Yishi Piao and Yixuan Chen and Yong Xiong and Yuchen Xuan and Yuehan Yang and Yuer Xu and Yukun Zha and Yunxian Ma and Yuping Lin and Yuting Yan and Yutong Xie and Yuwen Sheng and Yuxuan Zhu and Zekai Zhang and Zhe Ju and Zhenzhen Lin and Zheren Gao and Zheyang Sun and Zhigang Yan and Zhongyu Wu and Zi Wang and Zihua Qu and Ziling Yan and Ziyi Wan},
      year={2026},
      eprint={2609.19969},
      archivePrefix={arXiv},
      primaryClass={cs.CL},
      url={https://arxiv.org/abs/2609.19969}, 
}

@inproceedings{
    bai2025diep,
    title={Di{EP}: Adaptive Mixture-of-Experts Compression through Differentiable Expert Pruning},
    author={Sikai Bai and Haoxi Li and Jie ZHANG and Zicong Hong and Song Guo},
    booktitle={The Thirty-ninth Annual Conference on Neural Information Processing Systems},
    year={2025},
    url={https://openreview.net/forum?id=oTocSpfYvJ}
}

@misc{dong2025domainspecific,
      title={Domain-Specific Pruning of Large Mixture-of-Experts Models with Few-shot Demonstrations}, 
      author={Zican Dong and Han Peng and Peiyu Liu and Wayne Xin Zhao and Dong Wu and Feng Xiao and Zhifeng Wang},
      year={2025},
      eprint={2504.06792},
      archivePrefix={arXiv},
      primaryClass={cs.CL},
      url={https://arxiv.org/abs/2504.06792}, 
}

@misc{liu2026evoesapnonuniformexpertpruning,
      title={EvoESAP: Non-Uniform Expert Pruning for Sparse MoE}, 
      author={Zongfang Liu and Shengkun Tang and Boyang Sun and Zhiqiang Shen and Xin Yuan},
      year={2026},
      eprint={2603.06003},
      archivePrefix={arXiv},
      primaryClass={cs.LG},
      url={https://arxiv.org/abs/2603.06003}, 
}

@inproceedings{lu-etal-2024-experts,
    title = "Not All Experts are Equal: Efficient Expert Pruning and Skipping for Mixture-of-Experts Large Language Models",
    author = "Lu, Xudong  and
      Liu, Qi  and
      Xu, Yuhui  and
      Zhou, Aojun  and
      Huang, Siyuan  and
      Zhang, Bo  and
      Yan, Junchi  and
      Li, Hongsheng",
    editor = "Ku, Lun-Wei  and
      Martins, Andre  and
      Srikumar, Vivek",
    booktitle = "Proceedings of the 62nd Annual Meeting of the {A}ssociation for {C}omputational {L}inguistics (Volume 1: Long Papers)",
    month = aug,
    year = "2024",
    address = "Bangkok, Thailand",
    publisher = "Association for Computational Linguistics",
    url = "https://aclanthology.org/2024.acl-long.334/",
    doi = "10.18653/v1/2024.acl-long.334",
    pages = "6159--6172",
}

@inproceedings{
patil2025bfcl,
title={The Berkeley Function Calling Leaderboard ({BFCL}): From Tool Use to Agentic Evaluation of Large Language Models},
author={Shishir G Patil and Huanzhi Mao and Fanjia Yan and Charlie Cheng-Jie Ji and Vishnu Suresh and Ion Stoica and Joseph E. Gonzalez},
booktitle={Forty-second International Conference on Machine Learning},
year={2025},
url={https://openreview.net/forum?id=2GmDdhBdDk}
}

@article{platypus2023,
    title={Platypus: Quick, Cheap, and Powerful Refinement of LLMs}, 
    author={Ariel N. Lee and Cole J. Hunter and Nataniel Ruiz},
    booktitle={arXiv preprint arxiv:2308.07317},
    year={2023}
}

@inproceedings{
    luo2024wizardcoder,
    title={WizardCoder: Empowering Code Large Language Models with Evol-Instruct},
    author={Ziyang Luo and Can Xu and Pu Zhao and Qingfeng Sun and Xiubo Geng and Wenxiang Hu and Chongyang Tao and Jing Ma and Qingwei Lin and Daxin Jiang},
    booktitle={The Twelfth International Conference on Learning Representations},
    year={2024},
    url={https://openreview.net/forum?id=UnUwSIgK5W}
}

@inproceedings{
    lambert2025tulu,
    title={Tulu 3: Pushing Frontiers in Open Language Model Post-Training},
    author={Nathan Lambert and Jacob Morrison and Valentina Pyatkin and Shengyi Huang and Hamish Ivison and Faeze Brahman and Lester James Validad Miranda and Alisa Liu and Nouha Dziri and Xinxi Lyu and Yuling Gu and Saumya Malik and Victoria Graf and Jena D. Hwang and Jiangjiang Yang and Ronan Le Bras and Oyvind Tafjord and Christopher Wilhelm and Luca Soldaini and Noah A. Smith and Yizhong Wang and Pradeep Dasigi and Hannaneh Hajishirzi},
    booktitle={Second Conference on Language Modeling},
    year={2025},
    url={https://openreview.net/forum?id=i1uGbfHHpH}
}

@misc{penedo2024the,
      title={The FineWeb Datasets: Decanting the Web for the Finest Text Data at Scale}, 
      author={Guilherme Penedo and Hynek Kydlíček and Loubna Ben allal and Anton Lozhkov and Margaret Mitchell and Colin Raffel and Leandro Von Werra and Thomas Wolf},
      year={2024},
      eprint={2406.17557},
      archivePrefix={arXiv},
      primaryClass={cs.CL},
      url={https://arxiv.org/abs/2406.17557}, 
}

@misc{penedo2025codeforces,
      title={CodeForces CoTs}, 
      author={Guilherme Penedo and Anton Lozhkov and Hynek Kydlíček and Loubna Ben Allal and Edward Beeching and Agustín Piqueres Lajarín and Quentin Gallouédec and Nathan Habib and Lewis Tunstall and Leandro von Werra},
      year={2025},
      publisher = {Hugging Face},
      journal = {Hugging Face repository},
      howpublished = {\url{https://huggingface.co/datasets/open-r1/codeforces-cots}}
}

@inproceedings{
    liu2024apigen,
    title={{APIG}en: Automated {PI}peline for Generating Verifiable and Diverse Function-Calling Datasets},
    author={Zuxin Liu and Thai Quoc Hoang and Jianguo Zhang and Ming Zhu and Tian Lan and Shirley Kokane and Juntao Tan and Weiran Yao and Zhiwei Liu and Yihao Feng and Rithesh R N and Liangwei Yang and Silvio Savarese and Juan Carlos Niebles and Huan Wang and Shelby Heinecke and Caiming Xiong},
    booktitle={The Thirty-eight Conference on Neural Information Processing Systems Datasets and Benchmarks Track},
    year={2024},
    url={https://openreview.net/forum?id=Jfg3vw2bjx}
}

@inproceedings{
    yang2026swesmith,
    title={{SWE}-smith: Scaling Data for Software Engineering Agents},
    author={John Yang and Kilian Lieret and Carlos E Jimenez and Alexander Wettig and Kabir Khandpur and Yanzhe Zhang and Binyuan Hui and Ofir Press and Ludwig Schmidt and Diyi Yang},
    booktitle={The Thirty-ninth Annual Conference on Neural Information Processing Systems Datasets and Benchmarks Track},
    year={2026},
    url={https://openreview.net/forum?id=63iVrXc8cC}
}

@misc{pritsker2024writingprompts,
  author       = {Jade Pritsker},
  title        = {{WritingPrompts}: Curated Dataset},
  year         = {2024},
  month        = dec,
  publisher    = {Hugging Face},
  howpublished = {Hugging Face Datasets},
  url          = {https://huggingface.co/datasets/euclaise/WritingPrompts_curated}
}

@inproceedings{
    jimenez2024swebench,
    title={{SWE}-bench: Can Language Models Resolve Real-world Github Issues?},
    author={Carlos E Jimenez and John Yang and Alexander Wettig and Shunyu Yao and Kexin Pei and Ofir Press and Karthik R Narasimhan},
    booktitle={The Twelfth International Conference on Learning Representations},
    year={2024},
    url={https://openreview.net/forum?id=VTF8yNQM66}
}

@misc{guha2025openthoughtsdatarecipesreasoning,
  title={OpenThoughts: Data Recipes for Reasoning Models}, 
  author={Etash Guha and Ryan Marten and Sedrick Keh and Negin Raoof and Georgios Smyrnis and Hritik Bansal and Marianna Nezhurina and Jean Mercat and Trung Vu and Zayne Sprague and Ashima Suvarna and Benjamin Feuer and Liangyu Chen and Zaid Khan and Eric Frankel and Sachin Grover and Caroline Choi and Niklas Muennighoff and Shiye Su and Wanjia Zhao and John Yang and Shreyas Pimpalgaonkar and Kartik Sharma and Charlie Cheng-Jie Ji and Yichuan Deng and Sarah Pratt and Vivek Ramanujan and Jon Saad-Falcon and Jeffrey Li and Achal Dave and Alon Albalak and Kushal Arora and Blake Wulfe and Chinmay Hegde and Greg Durrett and Sewoong Oh and Mohit Bansal and Saadia Gabriel and Aditya Grover and Kai-Wei Chang and Vaishaal Shankar and Aaron Gokaslan and Mike A. Merrill and Tatsunori Hashimoto and Yejin Choi and Jenia Jitsev and Reinhard Heckel and Maheswaran Sathiamoorthy and Alexandros G. Dimakis and Ludwig Schmidt},
  year={2025},
  eprint={2506.04178},
  archivePrefix={arXiv},
  primaryClass={cs.LG},
  url={https://arxiv.org/abs/2506.04178}, 
}

@misc{park2026slimwisedecouplingexpertpruning,
      title={SlimWise: Decoupling Expert Pruning Across Prefill and Decode for Efficient MoE Serving}, 
      author={Gunho Park and Kyoungho Jeun and Juntaek Oh and Byeongjun Shin and Baeseong Park and Minsoo Rhu},
      year={2026},
      eprint={2609.34117},
      archivePrefix={arXiv},
      primaryClass={cs.LG},
      url={https://arxiv.org/abs/2609.34117}, 
}

@misc{tang2024hobbitmixedprecisionexpert,
      title={HOBBIT: A Mixed Precision Expert Offloading System for Fast MoE Inference}, 
      author={Peng Tang and Jiacheng Liu and Xiaofeng Hou and Yifei Pu and Jing Wang and Pheng-Ann Heng and Chao Li and Minyi Guo},
      year={2024},
      eprint={2411.01433},
      archivePrefix={arXiv},
      primaryClass={cs.LG},
      url={https://arxiv.org/abs/2411.01433}, 
}

@misc{yang2026freetokenefficientedgenativemoe,
      title={FreeToken: Efficient Edge-Native MoE Serving with Bandwidth-Adaptive Execution}, 
      author={Shuo Yang and Xiaoze Fan and Melissa Pan and Haocheng Xi and Zhe Wang and Shanlin Sun and Kurt Keutzer and Song Han and Matei Zaharia and Chenfeng Xu and Ion Stoica},
      year={2026},
      eprint={2608.16157},
      archivePrefix={arXiv},
      primaryClass={cs.DC},
      url={https://arxiv.org/abs/2608.16157}, 
}

@inproceedings{pruning-large-language,
title = {Instruction-Following Pruning for Large Language Models},
booktitle = {ICML},
author = {Bairu Hou and Qibin Chen and Jianyu Wang and Guoli Yin and Chong Wang and Nan Du and Ruoming Pang and Shiyu Chang and Tao Lei},
year = {2025},
URL = {https://arxiv.org/abs/2501.02086}
}

@misc{dettmers2026runtimedynamiccompression,
  author       = {Dettmers, Tim},
  title        = {Runtime Dynamic Compression of Mixture of Experts},
  year         = {2026},
  howpublished = {\url{https://timdettmers.com/papers/runtime-dynamic-compression.pdf}},
  note         = {Accessed: 2026-10-13}
}

@misc{austin2021programsynthesislargelanguage,
      title={Program Synthesis with Large Language Models}, 
      author={Jacob Austin and Augustus Odena and Maxwell Nye and Maarten Bosma and Henryk Michalewski and David Dohan and Ellen Jiang and Carrie Cai and Michael Terry and Quoc Le and Charles Sutton},
      year={2021},
      eprint={2108.07732},
      archivePrefix={arXiv},
      primaryClass={cs.PL},
      url={https://arxiv.org/abs/2108.07732}, 
}

@misc{shazeer2017outrageouslylargeneuralnetworks,
      title={Outrageously Large Neural Networks: The Sparsely-Gated Mixture-of-Experts Layer}, 
      author={Noam Shazeer and Azalia Mirhoseini and Krzysztof Maziarz and Andy Davis and Quoc Le and Geoffrey Hinton and Jeff Dean},
      year={2017},
      eprint={1701.06538},
      archivePrefix={arXiv},
      primaryClass={cs.LG},
      url={https://arxiv.org/abs/1701.06538}, 
}

@article{frantar-sparsegpt,
  title={{SparseGPT}: Massive Language Models Can Be Accurately Pruned in One-Shot}, 
  author={Elias Frantar and Dan Alistarh},
  year={2023},
  journal={arXiv preprint arXiv:2301.00774}
}

@inproceedings{
    lasby2026reap,
    title={{REAP} the Experts: Why Pruning Prevails for One-Shot MoE compression},
    author={Mike Lasby and Ivan Lazarevich and Nish Sinnadurai and Sean Lie and Yani Ioannou and Vithursan Thangarasa},
    booktitle={The Fourteenth International Conference on Learning Representations},
    year={2026},
    url={https://openreview.net/forum?id=ukGxWd2aDG}
}

@misc{xue2025moeinfinityefficientmoeinference,
      title={MoE-Infinity: Efficient MoE Inference on Personal Machines with Sparsity-Aware Expert Cache}, 
      author={Leyang Xue and Yao Fu and Zhan Lu and Luo Mai and Mahesh Marina},
      year={2025},
      eprint={2401.14361},
      archivePrefix={arXiv},
      primaryClass={cs.LG},
      url={https://arxiv.org/abs/2401.14361}, 
}

@misc{yang2026zipmoeefficientondevicemoe,
      title={ZipMoE: Efficient On-Device MoE Serving via Lossless Compression and Cache-Affinity Scheduling}, 
      author={Yuchen Yang and Yaru Zhao and Pu Yang and Shaowei Wang and Zhi-Hua Zhou},
      year={2026},
      eprint={2601.21198},
      archivePrefix={arXiv},
      primaryClass={cs.DC},
      url={https://arxiv.org/abs/2601.21198}, 
}

@inproceedings{10.1145/3669940.3707267,
    author = {Cao, Shiyi and Liu, Shu and Griggs, Tyler and Schafhalter, Peter and Liu, Xiaoxuan and Sheng, Ying and Gonzalez, Joseph E. and Zaharia, Matei and Stoica, Ion},
    title = {MoE-Lightning: High-Throughput MoE Inference on Memory-constrained GPUs},
    year = {2025},
    isbn = {9798400706981},
    publisher = {Association for Computing Machinery},
    address = {New York, NY, USA},
    url = {https://doi.org/10.1145/3669940.3707267},
    doi = {10.1145/3669940.3707267},
    booktitle = {Proceedings of the 30th ACM International Conference on Architectural Support for Programming Languages and Operating Systems, Volume 1},
    pages = {715–730},
    numpages = {16},
    location = {Rotterdam, Netherlands},
    series = {ASPLOS '25}
}

@misc{openai2025gptoss120bgptoss20bmodel,
      title={gpt-oss-120b \& gpt-oss-20b Model Card}, 
      author={OpenAI},
      year={2025},
      eprint={2508.10925},
      archivePrefix={arXiv},
      primaryClass={cs.CL},
      url={https://arxiv.org/abs/2508.10925}, 
}

@inproceedings{3618408.3619696,
    author = {Sheng, Ying and Zheng, Lianmin and Yuan, Binhang and Li, Zhuohan and Ryabinin, Max and Chen, Beidi and Liang, Percy and R{\'e}, Christopher and Stoica, Ion and Zhang, Ce},
    title = {FlexGen: high-throughput generative inference of large language models with a single GPU},
    year = {2023},
    publisher = {JMLR.org},
    booktitle = {Proceedings of the 40th International Conference on Machine Learning},
    articleno = {1288},
    numpages = {23},
    location = {Honolulu, Hawaii, USA},
    series = {ICML'23}
}

@inproceedings{3691938.3691945,
    author = {Agrawal, Amey and Kedia, Nitin and Panwar, Ashish and Mohan, Jayashree and Kwatra, Nipun and Gulavani, Bhargav S. and Tumanov, Alexey and Ramjee, Ramachandran},
    title = {Taming throughput-latency tradeoff in LLM inference with sarathi-serve},
    year = {2024},
    isbn = {978-1-939133-40-3},
    publisher = {USENIX Association},
    address = {USA},
    booktitle = {Proceedings of the 18th USENIX Conference on Operating Systems Design and Implementation},
    articleno = {7},
    numpages = {18},
    location = {Santa Clara, CA, USA},
    series = {OSDI'24}
}

@inproceedings{triton,
author = {Tillet, Philippe and Kung, H. T. and Cox, David},
title = {Triton: an intermediate language and compiler for tiled neural network computations},
year = {2019},
isbn = {9781450367196},
publisher = {Association for Computing Machinery},
address = {New York, NY, USA},
url = {https://doi.org/10.1145/3315508.3329973},
doi = {10.1145/3315508.3329973},
booktitle = {Proceedings of the 3rd ACM SIGPLAN International Workshop on Machine Learning and Programming Languages},
pages = {10–19},
numpages = {10},
location = {Phoenix, AZ, USA},
series = {MAPL 2019}
}

@article{chen2021codex,
  title={Evaluating Large Language Models Trained on Code},
  author={Mark Chen and Jerry Tworek and Heewoo Jun and Qiming Yuan and Henrique Ponde de Oliveira Pinto and Jared Kaplan and Harri Edwards and Yuri Burda and Nicholas Joseph and Greg Brockman and Alex Ray and Raul Puri and Gretchen Krueger and Michael Petrov and Heidy Khlaaf and Girish Sastry and Pamela Mishkin and Brooke Chan and Scott Gray and Nick Ryder and Mikhail Pavlov and Alethea Power and Lukasz Kaiser and Mohammad Bavarian and Clemens Winter and Philippe Tillet and Felipe Petroski Such and Dave Cummings and Matthias Plappert and Fotios Chantzis and Elizabeth Barnes and Ariel Herbert-Voss and William Hebgen Guss and Alex Nichol and Alex Paino and Nikolas Tezak and Jie Tang and Igor Babuschkin and Suchir Balaji and Shantanu Jain and William Saunders and Christopher Hesse and Andrew N. Carr and Jan Leike and Josh Achiam and Vedant Misra and Evan Morikawa and Alec Radford and Matthew Knight and Miles Brundage and Mira Murati and Katie Mayer and Peter Welinder and Bob McGrew and Dario Amodei and Sam McCandlish and Ilya Sutskever and Wojciech Zaremba},
  year={2021},
  eprint={2107.03374},
  archivePrefix={arXiv},
  primaryClass={cs.LG}
}

@misc{creative-writing-bench-v3,
  author = {Samuel J Paech},
  title = {EQ-Bench Creative Writing Benchmark v3},
  year = {2025},
  publisher = {GitHub},
  journal = {GitHub repository},
  howpublished = {\url{https://github.com/EQ-bench/creative-writing-bench}}
}

@misc{xuan2025mmluproxmultilingualbenchmarkadvanced,
      title={MMLU-ProX: A Multilingual Benchmark for Advanced Large Language Model Evaluation}, 
      author={Weihao Xuan and Rui Yang and Heli Qi and Qingcheng Zeng and Yunze Xiao and Aosong Feng and Dairui Liu and Yun Xing and Junjue Wang and Fan Gao and Jinghui Lu and Yuang Jiang and Huitao Li and Xin Li and Kunyu Yu and Ruihai Dong and Shangding Gu and Yuekang Li and Xiaofei Xie and Felix Juefei-Xu and Foutse Khomh and Osamu Yoshie and Qingyu Chen and Douglas Teodoro and Nan Liu and Randy Goebel and Lei Ma and Edison Marrese-Taylor and Shijian Lu and Yusuke Iwasawa and Yutaka Matsuo and Irene Li},
      year={2025},
      eprint={2503.10497},
      archivePrefix={arXiv},
      primaryClass={cs.CL},
      url={https://arxiv.org/abs/2503.10497}, 
}

@software{gerganov_llamacpp_2026,
	title = {llama.cpp},
	url = {https://github.com/ggml-org/llama.cpp},
	publisher = {ggml},
	author = {Gerganov, Georgi},
	month = sep,
	year = {2026},
}

@misc{panferov2026disaggregatedquantizationspecializingllm,
      title={Disaggregated Quantization: Specializing LLM Prefill and Decode}, 
      author={Andrei Panferov and Maximilian Kleinegger and Sweta Priyadarshi and Tijmen Blankevoort and Dan Alistarh},
      year={2026},
      eprint={2609.26333},
      archivePrefix={arXiv},
      primaryClass={cs.LG},
      url={https://arxiv.org/abs/2609.26333}, 
}

@misc{jain2024livecodebench,
      title={LiveCodeBench: Holistic and Contamination Free Evaluation of Large Language Models for Code}, 
      author={Naman Jain and King Han and Alex Gu and Wen-Ding Li and Fanjia Yan and Tianjun Zhang and Sida Wang and Armando Solar-Lezama and Koushik Sen and Ion Stoica},
      year={2024},
      eprint={2403.07974},
      archivePrefix={arXiv},
      primaryClass={cs.SE},
      url={https://arxiv.org/abs/2403.07974}, 
}

@inproceedings{
sieberling2025evopress,
title={EvoPress: Accurate Dynamic Model Compression via Evolutionary Search},
author={Oliver Sieberling and Denis Kuznedelev and Eldar Kurtic and Dan Alistarh},
booktitle={Forty-second International Conference on Machine Learning},
year={2025},
url={https://openreview.net/forum?id=l7QzcZpjc5}
}

@article{wang2025d,
  title={D\textsuperscript{2}MoE: Dual Routing and Dynamic Scheduling for Efficient On-Device MoE-based LLM Serving},
  author={Wang, Haodong and Zhou, Qihua and Hong, Zicong and Guo, Song},
  journal={arXiv preprint arXiv:2504.15299},
  year={2025}
}

@article{gptq,
  title={{GPTQ}: Accurate post-training quantization for generative pre-trained transformers},
  author={Frantar, Elias and Ashkboos, Saleh and Hoefler, Torsten and Alistarh, Dan},
  journal={arXiv preprint arXiv:2210.17323},
  year={2022}
}

@article{AQLM,
  title={Extreme Compression of Large Language Models via Additive Quantization},
  author={Egiazarian, Vage and Panferov, Andrei and Kuznedelev, Denis and Frantar, Elias and Babenko, Artem and Alistarh, Dan},
  journal={arXiv preprint arXiv:2401.06118},
  year={2024}
}

@misc{Tseng2024_QTIP,
      title={QTIP: Quantization with Trellises and Incoherence Processing}, 
      author={Albert Tseng and Qingyao Sun and David Hou and Christopher De Sa},
      year={2025},
      eprint={2406.11235},
      archivePrefix={arXiv},
      primaryClass={cs.LG},
      url={https://arxiv.org/abs/2406.11235}, 
}

@software{eval-harness,
  author       = {Gao, Leo and
                  Tow, Jonathan and
                  Biderman, Stella and
                  Black, Sid and
                  DiPofi, Anthony and
                  Foster, Charles and
                  Golding, Laurence and
                  Hsu, Jeffrey and
                  McDonell, Kyle and
                  Muennighoff, Niklas and
                  Phang, Jason and
                  Reynolds, Laria and
                  Tang, Eric and
                  Thite, Anish and
                  Wang, Ben and
                  Wang, Kevin and
                  Zou, Andy},
  title        = {A framework for few-shot language model evaluation},
  month        = sep,
  year         = 2021,
  publisher    = {Zenodo},
  version      = {v0.0.1},
  doi          = {10.5281/zenodo.5371628},
  url          = {https://doi.org/10.5281/zenodo.5371628},
}

@article{lin2024awq,
  title={AWQ: Activation-aware Weight Quantization for On-Device LLM Compression and Acceleration},
  author={Lin, Ji and Tang, Jiaming and Tang, Haotian and Yang, Shang and Chen, Wei-Ming and Wang, Wei-Chen and Xiao, Guangxuan and Dang, Xingyu and Gan, Chuang and Han, Song},
  journal={Proceedings of Machine Learning and Systems},
  volume={6},
  pages={87--100},
  year={2024}
}

@misc{hendrycks2021measuringmathematicalproblemsolving,
  title={Measuring Mathematical Problem Solving With the MATH Dataset},
  author={Dan Hendrycks and Collin Burns and Saurav Kadavath and Akul Arora and Steven Basart and Eric Tang and Dawn Song and Jacob Steinhardt},
  year={2021},
  eprint={2103.03874},
  archivePrefix={arXiv},
  primaryClass={cs.LG},
  url={https://arxiv.org/abs/2103.03874},
}

@misc{wang2024mmluprorobustchallengingmultitask,
  title={MMLU-Pro: A More Robust and Challenging Multi-Task Language Understanding Benchmark},
  author={Yubo Wang and Xueguang Ma and Ge Zhang and Yuansheng Ni and Abhranil Chandra and Shiguang Guo and Weiming Ren and Aaran Arulraj and Xuan He and Ziyan Jiang and Tianle Li and Max Ku and Kai Wang and Alex Zhuang and Rongqi Fan and Xiang Yue and Wenhu Chen},
  year={2024},
  eprint={2406.01574},
  archivePrefix={arXiv},
  primaryClass={cs.CL},
  url={https://arxiv.org/abs/2406.01574},
}

@misc{tseng2024quipbetterllmquantization,
      title={QuIP\#: Even Better LLM Quantization with Hadamard Incoherence and Lattice Codebooks}, 
      author={Albert Tseng and Jerry Chee and Qingyao Sun and Volodymyr Kuleshov and Christopher De Sa},
      year={2024},
      eprint={2402.04396},
      archivePrefix={arXiv},
      primaryClass={cs.LG},
      url={https://arxiv.org/abs/2402.04396}, 
}

@string{ICML = "International Conference on Machine Learning (ICML)"}

@string{SPRINGER = "Springer"}

@misc{wanda,
      title={A Simple and Effective Pruning Approach for Large Language Models}, 
      author={Mingjie Sun and Zhuang Liu and Anna Bair and J. Zico Kolter},
      year={2024},
      eprint={2306.11695},
      archivePrefix={arXiv},
      primaryClass={cs.CL}
}

@inproceedings{rein2024gpqa,
  title={Gpqa: A graduate-level google-proof q\&a benchmark},
  author={Rein, David and Hou, Betty Li and Stickland, Asa Cooper and Petty, Jackson and Pang, Richard Yuanzhe and Dirani, Julien and Michael, Julian and Bowman, Samuel R},
  booktitle={First Conference on Language Modeling},
  year={2024}
}

@inproceedings{liu2023dejavu,
  title     = {Deja Vu: Contextual Sparsity for Efficient {LLMs} at Inference Time},
  author    = {Liu, Zichang and Wang, Jue and Dao, Tri and Zhou, Tianyi and Yuan, Binhang and Song, Zhao and Shrivastava, Anshumali and Zhang, Ce and Tian, Yuandong and R{\'e}, Christopher and Chen, Beidi},
  booktitle = {International Conference on Machine Learning (ICML)},
  pages     = {22137--22176},
  year      = {2023}
}

@inproceedings{pei2025premoe,
  title     = {{PreMoE}: Proactive Inference for Efficient Mixture-of-Experts},
  author    = {Pei, Zehua and Zhang, Ying and Zhen, Hui-Ling and Yuan, Tao and Yu, Xianzhi and Dong, Zhenhua and Pan, Sinno Jialin and Yuan, Mingxuan and Yu, Bei},
  booktitle = {Conference on Language Modeling (COLM)},
  year      = {2026},
  url       = {https://arxiv.org/abs/2505.17639}
}

@misc{koo2026dep,
  title         = {Dynamic Expert Pruning for Multi-Agent Systems},
  author        = {Koo, Jabin and Abbasloo, Soheil and Lee, Sungjae and Ok, Jungseul},
  year          = {2026},
  eprint        = {2610.02951},
  archivePrefix = {arXiv},
  url           = {https://arxiv.org/abs/2610.02951}
}

@inproceedings{zhang2025daop,
  title     = {{DAOP}: Data-Aware Offloading and Predictive Pre-Calculation for Efficient {MoE} Inference},
  author    = {Zhang, Yujie and Aggarwal, Shivam and Mitra, Tulika},
  booktitle = {Design, Automation \& Test in Europe Conference (DATE)},
  pages     = {1--7},
  year      = {2025},
  doi       = {10.23919/DATE64628.2025.10992741}
}

@inproceedings{wang2026casmoe,
  title     = {{CasMoE}: A Cascaded Framework for Efficient {MoE} Inference on Resource-constrained Devices},
  author    = {Wang, Chengcheng and He, Haowen and Zhao, Liang and Deng, Xiaoheng and Duan, Lixin and Wan, Shaohua},
  booktitle = {Proceedings of the AAAI Conference on Artificial Intelligence},
  volume    = {40},
  number    = {31},
  pages     = {26133--26141},
  year      = {2026},
  doi       = {10.1609/aaai.v40i31.39816}
}

@inproceedings{mu2026domoe,
  title     = {{DoMoE}: Domain-Aware Semantic Expert Prediction for Efficient {MoE} Inference Under Expert Offloading},
  author    = {Mu, Yao and Chen, Fahao and Zhu, Wenbin and Zhao, Mengying and Shen, Zhaoyan and Yu, Dongxiao},
  booktitle = {Proceedings of the International Joint Conference on Artificial Intelligence (IJCAI)},
  pages     = {5901--5909},
  year      = {2026},
  doi       = {10.24963/ijcai.2026/657}
}

@misc{madan2026speculating,
  title         = {Speculating Experts Accelerates Inference for Mixture-of-Experts},
  author        = {Madan, Vivan and Singhania, Prajwal and Bhatele, Abhinav and Goldstein, Tom and Panda, Ashwinee},
  year          = {2026},
  eprint        = {2603.19289},
  archivePrefix = {arXiv},
  url           = {https://arxiv.org/abs/2603.19289}
}

@misc{yang2026moecore,
      title={MoE-CORE: Coordinated Expert Offloading and Residency for Memory-Constrained MoE Inference}, 
      author={Ke Yang and Yongji Gao and Xushi Li and Kui Luo and Sicheng Zhang and Tianming Zhou and Keyi Liu and Shufang Lu and Aoxuan Chen and Jie Meng and Jingchun Gao and Dan Li and Xinkai You and Dan Li and Zhixiang Xia and Yan Shi and Yang Liu and Yanjia Zeng and Liangjun Feng},
      year={2026},
      eprint={2610.01950},
      archivePrefix={arXiv},
      primaryClass={cs.DC},
      url={https://arxiv.org/abs/2610.01950}, 
}

@inproceedings{zhou2025floe,
  title     = {{FloE}: On-the-Fly {MoE} Inference on Memory-constrained {GPU}},
  author    = {Zhou, Yuxin and Li, Zheng and Zhang, Jun and Wang, Jue and Wang, Yiping and Xie, Zhongle and Chen, Ke and Shou, Lidan},
  booktitle = {International Conference on Machine Learning (ICML)},
  pages     = {78859--78882},
  year      = {2025}
}
\bibliographystyle{mlsys2025}

\subsection*{AI Usage Disclosure}

In this work, we used generative AI tools, particularly Claude Code, to assist with prototyping and implementation, including the serving engine, evaluation pipelines, and benchmarking code. Claude Code was also used to iteratively improve source code and its quality. Furthermore, we used ChatGPT to improve the clarity and readability of manuscript text initially written by the authors.

All AI-assisted code was reviewed, tested, and validated through experiments and evaluations. Numerical claims were checked against experimental results, and all AI-assisted text was reviewed and revised by the authors. We take full responsibility for the final content of this work, including all text, claims, code, and artifacts produced with the assistance of generative AI.

\onecolumn
\appendix

\newpage

\section{Limitations \& Future Work} 
Even though KEEP pushes beyond the 50\% sparsity barrier, our system is still limited to single-request inference, and our evaluation to only a few models. Therefore, future work could explore context-aware batching, as important expert sets within similar domains often overlap, allowing requests to share experts. Furthermore, we only explored dynamic sparsification with a fixed expert set per request, which could instead be adapted during decoding. Lastly, we only generated up to 64k tokens, and longer generations could worsen our results, as they might gradually diverge from their original task, especially at high sparsity levels. 
Finally, an agentic system could combine context-aware batching with splitting long tasks into shorter, more focused requests, potentially improving efficiency and accuracy.

\section{Expert Contribution, Selection \& Prediction Ablation}
\label{sec:expert_pred_ablation}

\begin{figure*}[h]
    \centering
    \includegraphics[width=1\linewidth]{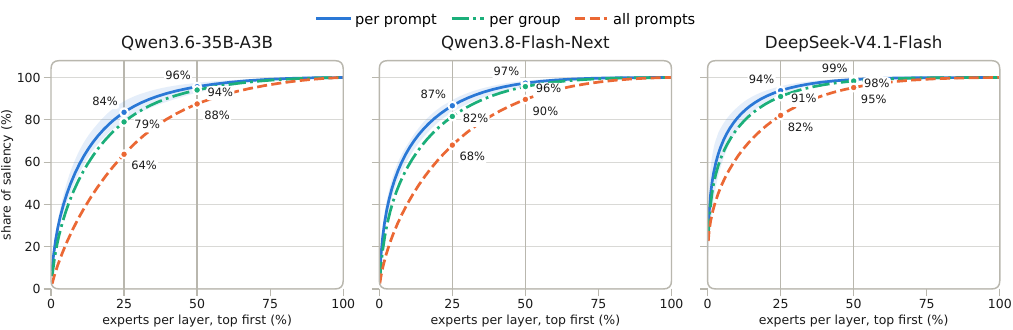}
    \caption{Share of total saliency captured by the top-\% highest-saliency experts of each MoE layer, across 24 prompts in 8 task groups (mean over layers). Experts are ranked by saliency \emph{per prompt} (each prompt has its own ranking, mean over prompts), \emph{per group} (one ranking shared across prompts within a group, mean over groups), or \emph{across all prompts} (one ranking shared across all prompts). Dots mark the contribution of the top-25\% and top-50\% most important experts. Experts per layer: 256, 512, 384.}
    \label{fig:contribution_ablation}
\end{figure*}

\begin{figure*}[h]
    \centering
    \includegraphics[width=1\linewidth]{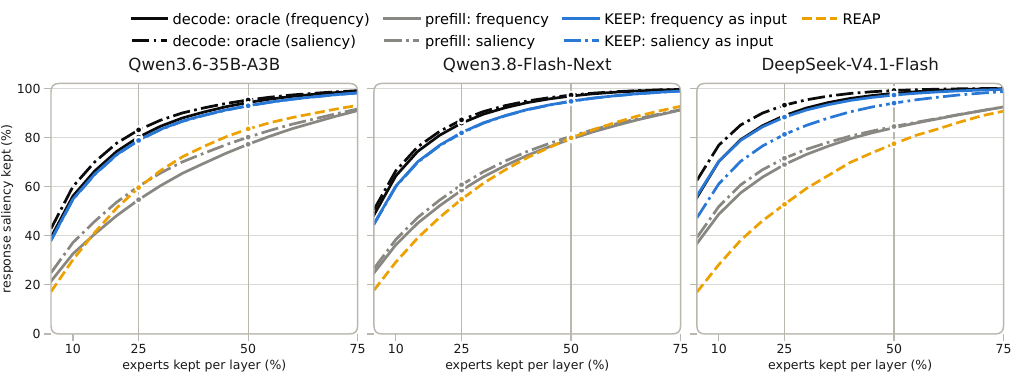}
    \caption{Ablation of expert contribution: share of the response's saliency captured by the top-\% of experts per layer, selected from the prefill or the decode (hindsight) based on frequency or saliency. We compare them with REAP and KEEP, with either frequency or saliency as its input.}
    \label{fig:saliency_ablation}
\end{figure*}

\textbf{Setup.} To evaluate expert contribution patters for expert selection and prediction, we compute this saliency score for Qwen3.6-35B-A3B~\cite{qwen3.6-35b-a3b}, Qwen3.8-Flash-Next~\cite{qiu2026designqwen38nextarchitectureevaluation} and DeepSeek-V4.1-Flash~\cite{deepseekai2026deepseekv41flashpushinglimitskv} for a set of randomly picked prompts from random domains. We select randomly 3 prompts from Math500~\cite{hendrycks2021measuringmathematicalproblemsolving}, LiveCodeBench v6~\cite{jain2024livecodebench}, GPQA-Diamond~\cite{rein2024gpqa}, Creative Writing Bench~\cite{creative-writing-bench-v3}, as well as 3 prompts from MMLU-Pro(X)~\cite{wang2024mmluprorobustchallengingmultitask, xuan2025mmluproxmultilingualbenchmarkadvanced} for the languages English, German, Chinese and French.

\textbf{Expert Contribution.} Figure \ref{fig:contribution_ablation} provides a more detailed view of Figure \ref{fig:expert_frequency}, showing how the retained expert contribution decreases with increasing sparsity when experts are selected per prompt, per domain, or globally across domains.

\textbf{Expert Selection.} After our initial analysis of expert saliency concentration, we wanted to determine the best set of experts to use. For this, we used our generated responses and compared different sets of experts based on their overall contribution. This allowed us to efficiently establish an upper bound and evaluate how closely our method could approach it. 
In Figure~\ref{fig:saliency_ablation}, we can see that the best set of experts is obtained by considering the full decoding sequence in hindsight. We take this as our oracle, as it allows us to select experts based on information from the entire recorded sequence, which would not be available during actual decoding. This is our upper bound for expert selection on these traces and we evaluate how closely we can approach it.
Therefore, we first observe that selecting experts based on saliency, either during prefill or in hindsight of decoding, provides a stronger candidate than a frequency-based selection. However, we can see that the expert set determined during prefill performs rather poorly, and we would instead need the set determined in hindsight of decoding.

\textbf{Expert Prediction.} If we compare against REAP, we can see that it either performs on par with the set determined during prefill or even worse, showcasing its limited generalization. However, if we use the set determined by KEEP, we essentially close the gap to the oracle and perform well. Furthermore, we see that using saliency as an input to KEEP does not improve its performance and potentially even hurts it slightly. For this reason, there is no need for this more expensive computation during prefill.

\textbf{Limitation.} We only consider the decoding expert set, as we establish that performing prefill with a pruned model leads to worse accuracy (see Appendix~\ref{sec:pruned_prefill}), while the router outputs during prefill already provide the information needed to determine expert importance. We, also want to acknowledge that this study is based only on already computed traces. Therefore, our oracle represents an upper bound for expert selection on these traces, but not necessarily for downstream accuracy, as changing the expert set could also change the generated responses and subsequent expert activations. Evaluating this directly would require running all downstream tasks, recording and computing their saliency, and then rerunning each task with the best expert set determined by the first run. Since this would have been too expensive, we opted for this faster approximation to estimate how well the different expert selection methods could perform.

\section{Experiment Design}
\label{sec:exp_design}

\subsection{Calibration Data}
\label{section:calibration_data}

To calibrate KEEP and REAP, we use a mix of Open-Platypus~\cite{platypus2023}, OpenThoughts-114k~\cite{guha2025openthoughtsdatarecipesreasoning}, evol-codealpaca~\cite{luo2024wizardcoder}, tulu-3-sft-personas-math~\cite{lambert2025tulu} and FineWeb-Edu~\cite{penedo2024the} for general and domain-specific calibration. For long coding tasks, we add codeforces-cots~\cite{penedo2025codeforces}, for tool calling and agentic coding, xlam-function-calling~\cite{liu2024apigen} and SWE-smith-trajectories~\cite{yang2026swesmith} respectively, and for creative writing, WritingPrompts curated~\cite{pritsker2024writingprompts}. KEEP generated all the responses itself, while REAP used the given prompt-response pairs. The exact composition, generation budget and whether thinking was allow can be seen in Table~\ref{tab:hp-calibration}. As we see, we try to cover, long and short generation aswell as thinking and non-thinking generation.

\begin{table*}[h]
    \centering\small
    \caption{Calibration mix. Budget: maximum number of new tokens per response.}
    \label{tab:hp-calibration}
    \begin{tabular}{llrrc}
    \toprule
    Domain & Source & Prompts & Budget & Thinking \\
    \midrule
    \multirow{4}{*}{General} & Open-Platypus & 256 & 32k & on \\
     & OpenThoughts-114k & 256 & 32k & on \\
     & evol-codealpaca-v1 & 256 & 32k & on \\
     & FineWeb-Edu & 256 & 32k & on \\
    \midrule
    Long reasoning & OpenThoughts-114k (longest 15\% by reference answer) & 384 & 64k & on \\
    Long code & OpenThoughts-114k, code problems (longest 25\%) & 384 & 64k & on \\
    Code & OpenThoughts-114k, code problems (longest 25\%) & 384 & 64k & off \\
    Competitive programming & Codeforces-CoTs (longest 25\%) & 384 & 64k & on \\
    Agentic coding & SWE-smith trajectories & 256 & 32k & on \\
    Function calling & xLAM function-calling-60k & 256 & 32k & on \\
    Creative writing & WritingPrompts (curated) & 256 & 32k & on \\
    Math (personas) & Tülu-3 SFT personas math & 256 & 32k & on \\
    \midrule
    Total & & 3,584 & & \\
    \bottomrule
    \end{tabular}
\end{table*}

For REAP~\citep{lasby2026reap}, we use the authors' reference implementation and its default hyperparameters. The saliency, is computed on our general calibration mix with sequence length of 2,048 from the prompts of our general calibration mix together with the datasets' reference answers, using default parameters. We also calculated it on the self-generated responses, but it performed worse, so we used their approach.

\subsection{Hyperparameters}

\begin{table*}[h]
    \centering\small
    \caption{Comparison of MoE models included in our study.}
    \label{tab:models}
    \begin{tabular}{lcccccc}
    \toprule
    Model & Routed Experts & Shared Experts & Top-k & Sparsity & Parameters (1e9) & Active Params. (1e9) \\
    \midrule
    Qwen3.6-35B-A3B & 256 & 1 & 8 & 96.50\% & 34.7 & 3.5 \\
    Qwen3.8-Flash-Next-FP8 & 512 & 1 & 10 & 97.86\% & 176.9 & 7.3 \\
    DeepSeek-V4.1-Flash & 384 & 1 & 6 & 98.18\% & 748.9 & 17.2 \\
    \bottomrule
    \end{tabular}
\end{table*}

This section presents the hyperparameters used for evaluation.

\textbf{Models.} We use models from their original providers, with the exception of the NVFP4 models, for which we use the NVIDIA versions available on Hugging Face. We use the models' default sampling settings, i.e., a temperature of 1.0 and a top-$p$ of 0.95, as well as a top-$k$ of 20 for the Qwen models. All models use thinking, with the reasoning effort set to \texttt{xhigh} for Qwen3.8-Flash-Next and \texttt{max} for DeepSeek-V4.1-Flash. The exact model properties can be found in Table~\ref{tab:models}.

\textbf{Benchmarks.} We evaluate using lm-evaluation-harness~\cite{eval-harness}, zero-shot with the chat template and on the full test sets: MATH-500~\cite{hendrycks2021measuringmathematicalproblemsolving} (500 problems, accuracy evaluated with math-verify), GPQA-Diamond~\cite{rein2024gpqa} (198 questions), HumanEval+~\cite{chen2021codex, evalplus}., (164 problems), and MBPP+~\cite{austin2021programsynthesislargelanguage, evalplus} (378 problems). For the latter two, we report pass@1. All four benchmarks run with thinking and at most 65,536 new tokens, while MMLU-Pro~\cite{wang2024mmluprorobustchallengingmultitask} (12,032 questions) runs without thinking and with at most 16,384 new tokens. Every benchmark runs with three seeds.

For SWE-bench Verified~\citep{jimenez2024swebench}, we use mini-swe-agent~\cite{yang2024sweagent} with tool calling on all 500 instances. The agent has a limit of 250 steps and no cost limit, a context length of 163,840 tokens, at most 32,768 new tokens per call, and a timeout of 60 seconds per command. For BFCL-v4~\citep{patil2025bfcl}, we use bfcl-eval on its 17 non-agentic categories with 4,441 test cases, with thinking and at most 8,192 new tokens per call, and report accuracy. We also run this benchmark with three seeds, even though the temperature is close to 1.

\subsection{Hardware}

Table~\ref{tab:system} summarizes the hardware configurations used in our experiments, including measured PCIe transfer and disk read bandwidths.

\begin{table}[t]
  \centering
  \caption{Test systems. $B_P$ is the measured host-to-device
  expert-transfer bandwidth over PCIe and $B_D$ is the measured
  disk read bandwidth using \texttt{O\_DIRECT}.}
  \label{tab:system}
  \setlength{\tabcolsep}{3pt}
  \begin{tabular}{@{}llrrlrr@{}}
    \toprule
    System & GPU & VRAM (GB) & $B_P$ (GB/s)
      & CPU & DRAM (GiB) & $B_D$ (GB/s) \\
    \midrule
    5090 desktop
      & RTX 5090 & 32 & 27.9
      & i9-14900K (32 threads) & 62 DDR5 & 6.7 \\
    3090 server
      & RTX 3090 & 24 & 25.8
      & 2$\times$ EPYC 7502 (128 threads) & 504 DDR4 & 6.9 \\
    B200 server
      & 8$\times$ B200 & 180/GPU & 53.7
      & 2$\times$ EPYC 9965 (768 threads) & 3,022 DDR5 & 26.2 \\
    \bottomrule
  \end{tabular}
\end{table}

\subsection{End-to-end Benchmarks}
\label{sec:hw_bench}

We serve the respective models one request at a time with KEEP at 0\%, 50\% and 75\% sparsity, FreeToken and llama.cpp. We measure time-to-first-token (TTFT) for prompts of 1k--16k tokens and decode throughput for outputs of 1k--16k tokens. Additionally, we run a four-part reasoning task turn by turn. Latency points are medians over four runs after a warm-up.

\section{Detailed Results}
\label{sec:appendix_results}

\begin{table*}[t]
  \caption{Detailed accuracy over MATH-500, GPQA-Diamond, MMLU-Pro, HumanEval+ and MBPP+ at 50\% and 75\% sparsity, for KEEP, REAP and SlimWise. Results are the mean of three seeds with standard deviation right next to it.}
  \label{tab:full}
  \centering
  \small
  \begin{tabular}{lllrrrrrr}
    \toprule
     &  &  & \multicolumn{2}{c}{Reasoning} & \multicolumn{1}{c}{Knowledge} & \multicolumn{2}{c}{Code} &  \\
    \cmidrule(lr){4-5} \cmidrule(lr){6-6} \cmidrule(lr){7-8}
    Model & Sparsity & Method & MATH-500 & GPQA-D & MMLU-Pro & HumanEval+ & MBPP+ & Mean \\
    \midrule
    Qwen3.6-35B-A3B & 0\% & dense & $88.1_{\pm 0.6}$ & $81.6_{\pm 1.2}$ & $83.8_{\pm 0.2}$ & $93.7_{\pm 0.4}$ & $83.6_{\pm 0.3}$ & $86.2_{\pm 0.2}$ \\
    \cmidrule(lr){2-9}
    \footnotesize (BF16) & \multirow{3}{*}{50\%} & REAP & $87.7_{\pm 1.6}$ & $61.6_{\pm 1.5}$ & $70.8_{\pm 0.2}$ & $92.1_{\pm 2.1}$ & $80.6_{\pm 1.1}$ & $78.5_{\pm 0.5}$ \\
     &  & SlimWise & $87.5_{\pm 2.0}$ & $69.0_{\pm 2.0}$ & $78.6_{\pm 0.1}$ & $93.7_{\pm 0.7}$ & $81.7_{\pm 0.8}$ & $82.1_{\pm 0.7}$ \\
     &  & KEEP & $86.0_{\pm 0.5}$ & $79.3_{\pm 3.1}$ & $83.5_{\pm 0.1}$ & $93.3_{\pm 0.6}$ & $83.2_{\pm 0.7}$ & $\mathbf{85.1}_{\pm 0.8}$ \\
    \cmidrule(lr){2-9}
     & \multirow{3}{*}{75\%} & REAP & $81.5_{\pm 0.9}$ & $29.5_{\pm 1.8}$ & $39.4_{\pm 0.4}$ & $77.4_{\pm 1.8}$ & $66.0_{\pm 1.5}$ & $58.8_{\pm 0.4}$ \\
     &  & SlimWise & $86.2_{\pm 0.7}$ & $34.0_{\pm 5.6}$ & $63.4_{\pm 0.2}$ & $85.4_{\pm 1.6}$ & $70.8_{\pm 0.6}$ & $68.0_{\pm 0.9}$ \\
     &  & KEEP & $81.3_{\pm 0.9}$ & $69.9_{\pm 1.2}$ & $78.4_{\pm 0.1}$ & $83.3_{\pm 3.1}$ & $75.8_{\pm 1.6}$ & $\mathbf{77.7}_{\pm 0.9}$ \\
    \midrule
    Qwen3.8-Flash-Next & 0\% & dense & $96.6_{\pm 0.2}$ & $82.2_{\pm 1.1}$ & $86.0_{\pm 0.1}$ & $95.1_{\pm 0.6}$ & $85.4_{\pm 0.4}$ & $89.1_{\pm 0.4}$ \\
    \cmidrule(lr){2-9}
    \footnotesize (FP8) & \multirow{3}{*}{50\%} & REAP & $96.5_{\pm 0.2}$ & $52.0_{\pm 2.2}$ & $74.4_{\pm 0.4}$ & $94.1_{\pm 0.7}$ & $83.3_{\pm 1.1}$ & $80.1_{\pm 0.6}$ \\
     &  & SlimWise & $97.3_{\pm 0.2}$ & $63.1_{\pm 2.6}$ & $82.7_{\pm 0.3}$ & $94.5_{\pm 0.6}$ & $84.4_{\pm 0.5}$ & $84.4_{\pm 0.4}$ \\
     &  & KEEP & $96.3_{\pm 0.1}$ & $81.0_{\pm 1.9}$ & $85.8_{\pm 0.1}$ & $94.7_{\pm 0.4}$ & $84.6_{\pm 1.0}$ & $\mathbf{88.5}_{\pm 0.5}$ \\
    \cmidrule(lr){2-9}
     & \multirow{3}{*}{75\%} & REAP & $90.8_{\pm 0.5}$ & $23.6_{\pm 0.8}$ & $24.7_{\pm 0.2}$ & $87.6_{\pm 0.9}$ & $77.7_{\pm 0.9}$ & $60.9_{\pm 0.2}$ \\
     &  & SlimWise & $94.2_{\pm 0.2}$ & $27.1_{\pm 3.2}$ & $51.9_{\pm 0.6}$ & $91.1_{\pm 2.1}$ & $79.8_{\pm 1.8}$ & $68.8_{\pm 1.0}$ \\
     &  & KEEP & $96.1_{\pm 0.5}$ & $80.1_{\pm 1.2}$ & $79.4_{\pm 0.2}$ & $92.7_{\pm 0.6}$ & $82.8_{\pm 2.0}$ & $\mathbf{86.2}_{\pm 0.6}$ \\
    \midrule
    DeepSeek-V4.1-Flash & 0\% & dense & $95.4_{\pm 0.4}$ & $88.6_{\pm 0.8}$ & $84.7_{\pm 0.0}$ & $95.1_{\pm 0.0}$ & $86.1_{\pm 0.4}$ & $90.0_{\pm 0.1}$ \\
    \cmidrule(lr){2-9}
    \footnotesize (MXFP4) & \multirow{3}{*}{50\%} & REAP & $95.7_{\pm 0.3}$ & $66.7_{\pm 1.0}$ & $80.3_{\pm 0.2}$ & $92.9_{\pm 0.9}$ & $84.3_{\pm 0.4}$ & $84.0_{\pm 0.3}$ \\
     &  & SlimWise & $96.1_{\pm 1.0}$ & $70.5_{\pm 0.3}$ & $83.8_{\pm 0.2}$ & $92.9_{\pm 0.4}$ & $85.5_{\pm 0.6}$ & $85.8_{\pm 0.3}$ \\
     &  & KEEP & $96.5_{\pm 0.5}$ & $88.7_{\pm 1.3}$ & $84.9_{\pm 0.2}$ & $94.3_{\pm 0.4}$ & $85.5_{\pm 0.4}$ & $\mathbf{90.0}_{\pm 0.3}$ \\
    \cmidrule(lr){2-9}
     & \multirow{3}{*}{75\%} & REAP & $79.5_{\pm 0.9}$ & $31.3_{\pm 0.9}$ & $53.7_{\pm 0.4}$ & $50.0_{\pm 3.0}$ & $50.3_{\pm 3.7}$ & $53.0_{\pm 1.0}$ \\
     &  & SlimWise & $80.1_{\pm 1.5}$ & $38.9_{\pm 4.6}$ & $70.2_{\pm 0.4}$ & $43.9_{\pm 3.7}$ & $49.7_{\pm 1.4}$ & $56.6_{\pm 1.3}$ \\
     &  & KEEP & $95.6_{\pm 0.4}$ & $86.9_{\pm 1.8}$ & $84.2_{\pm 0.1}$ & $88.2_{\pm 1.9}$ & $81.7_{\pm 0.5}$ & $\mathbf{87.3}_{\pm 0.5}$ \\
    \bottomrule
  \end{tabular}
\end{table*}

Table \ref{tab:full} presents detailed evaluation results of Qwen3.6-35B-A3B, Qwen3.8-Flash-Next and DeepSeek-V4.1-Flash with KEEP versus independently pruned REAP models and SlimWise approach. 

\subsection{Transferability across Quantization Schemes}

Since, we ran into memory limitations for our end-to-end experiments, we used more heavily quantized versions of our models. Therefore, we evaluated how well the transfer matrix $T$, generated using a higher-precision model, transfers to the same model at lower precision. The results in Table~\ref{fig:transfer} show that $T$ remains effective at lower precision. We believe this holds as long as quantization does not substantially change the routing behavior. However, for real-world deployments, we still recommend generating $T$ using the same model representation that will be deployed.

\begin{table*}[t]
  \caption{Transfer across quantization formats. The predictor $T$, trained on BF16 (Qwen3.6) or FP8 (Qwen3.8), is applied unchanged to NVFP4. Recovery differs by at most 1.2 percentage points across both models and sparsity levels. Accuracy (\%) is reported as mean and standard deviation over three seeds; $\Delta$Rec. denotes the difference in recovery relative to the original precision.}
  \label{tab:transfer}
  \centering
  \tiny
  \begin{tabular}{llllrrrrrrr}
    \toprule
     &  &  &  & \multicolumn{2}{c}{Reasoning} & \multicolumn{1}{c}{Knowledge} & \multicolumn{2}{c}{Code} &  &  \\
    \cmidrule(lr){5-6} \cmidrule(lr){7-7} \cmidrule(lr){8-9}
    Model & Weights & $T$ & Sparsity & MATH-500 & GPQA-D & MMLU-Pro & HumanEval+ & MBPP+ & Mean & $\Delta$Rec. \\
    \midrule
    \multirow{6}{*}{Qwen3.6-35B-A3B} & \multirow{3}{*}{BF16} & \multirow{3}{*}{fitted} & 0\% & $88.1_{\pm 0.6}$ & $81.6_{\pm 1.2}$ & $83.8_{\pm 0.2}$ & $93.7_{\pm 0.4}$ & $83.6_{\pm 0.3}$ & $86.2_{\pm 0.2}$ & -- \\
     &  &  & 50\% & $86.0_{\pm 0.5}$ & $79.3_{\pm 3.1}$ & $83.5_{\pm 0.1}$ & $93.3_{\pm 0.6}$ & $83.2_{\pm 0.7}$ & $85.1_{\pm 0.8}$ & -- \\
     &  &  & 75\% & $81.3_{\pm 0.9}$ & $69.9_{\pm 1.2}$ & $78.4_{\pm 0.1}$ & $83.3_{\pm 3.1}$ & $75.8_{\pm 1.6}$ & $77.7_{\pm 0.9}$ & -- \\
    \cmidrule(lr){2-11}
     & \multirow{3}{*}{NVFP4} & \multirow{3}{*}{from BF16} & 0\% & $89.3_{\pm 1.2}$ & $79.1_{\pm 0.3}$ & $83.1_{\pm 0.2}$ & $94.3_{\pm 0.4}$ & $81.1_{\pm 1.3}$ & $85.4_{\pm 0.2}$ & -- \\
     &  &  & 50\% & $86.5_{\pm 1.0}$ & $79.1_{\pm 1.3}$ & $83.0_{\pm 0.2}$ & $94.3_{\pm 0.4}$ & $82.0_{\pm 0.5}$ & $85.0_{\pm 0.2}$ & $+0.8$ \\
     &  &  & 75\% & $81.9_{\pm 1.2}$ & $69.2_{\pm 1.8}$ & $77.1_{\pm 0.0}$ & $81.7_{\pm 1.1}$ & $73.5_{\pm 1.2}$ & $76.7_{\pm 0.1}$ & $-0.4$ \\
    \midrule
    \multirow{6}{*}{Qwen3.8-Flash-Next} & \multirow{3}{*}{FP8} & \multirow{3}{*}{fitted} & 0\% & $96.6_{\pm 0.2}$ & $82.2_{\pm 1.1}$ & $86.0_{\pm 0.1}$ & $95.1_{\pm 0.6}$ & $85.4_{\pm 0.4}$ & $89.1_{\pm 0.4}$ & -- \\
     &  &  & 50\% & $96.3_{\pm 0.1}$ & $81.0_{\pm 1.9}$ & $85.8_{\pm 0.1}$ & $94.7_{\pm 0.4}$ & $84.6_{\pm 1.0}$ & $88.5_{\pm 0.5}$ & -- \\
     &  &  & 75\% & $96.1_{\pm 0.5}$ & $80.1_{\pm 1.2}$ & $79.4_{\pm 0.2}$ & $92.7_{\pm 0.6}$ & $82.8_{\pm 2.0}$ & $86.2_{\pm 0.6}$ & -- \\
    \cmidrule(lr){2-11}
     & \multirow{3}{*}{NVFP4} & \multirow{3}{*}{from FP8} & 0\% & $95.7_{\pm 0.9}$ & $83.0_{\pm 1.1}$ & $85.8_{\pm 0.1}$ & $95.1_{\pm 0.6}$ & $84.6_{\pm 0.4}$ & $88.9_{\pm 0.2}$ & -- \\
     &  &  & 50\% & $96.3_{\pm 0.6}$ & $81.1_{\pm 1.5}$ & $85.5_{\pm 0.2}$ & $94.7_{\pm 1.3}$ & $85.0_{\pm 0.6}$ & $88.5_{\pm 0.2}$ & $+0.3$ \\
     &  &  & 75\% & $96.6_{\pm 0.8}$ & $73.7_{\pm 2.7}$ & $79.3_{\pm 0.1}$ & $92.9_{\pm 2.0}$ & $82.3_{\pm 0.3}$ & $85.0_{\pm 0.8}$ & $-1.2$ \\
    \bottomrule
  \end{tabular}
\end{table*}

\section{Pruned Prefill}
\label{sec:pruned_prefill}

\begin{table*}[t]
  \caption{}
  \caption{Prefill ablation for KEEP on Qwen3.6-35B-A3B: the prompt is prefilled densely (\emph{dense}), again among the experts KEEP selected (\emph{selected}), or among the REAP checkpoint's experts (\emph{static}, \emph{unmasked $r$}): KEEP's input router weights come from a pass over all experts). Accuracy (\%) on the full test sets. Mean over three seeds with standard deviation is reported. $\Delta$ is the gap to KEEP with a dense prefill at the same sparsity.}
  \label{tab:prefill}
  \centering
  \tiny
  \begin{tabular}{lllrrrrrrr}
    \toprule
     &  &  & \multicolumn{2}{c}{Reasoning} & \multicolumn{1}{c}{Knowledge} & \multicolumn{2}{c}{Code} &  &  \\
    \cmidrule(lr){4-5} \cmidrule(lr){6-6} \cmidrule(lr){7-8}
    Sparsity & Method & Prefill & MATH-500 & GPQA-D & MMLU-Pro & HumanEval+ & MBPP+ & Mean & $\Delta$ \\
    \midrule
    0\% & dense & dense & $89.3_{\pm 1.2}$ & $79.1_{\pm 0.3}$ & $83.1_{\pm 0.2}$ & $94.3_{\pm 0.4}$ & $81.1_{\pm 1.3}$ & $85.4_{\pm 0.2}$ & -- \\
    \cmidrule(lr){1-10}
    \multirow{4}{*}{50\%} & KEEP & dense & $86.5_{\pm 1.0}$ & $79.1_{\pm 1.3}$ & $83.0_{\pm 0.2}$ & $94.3_{\pm 0.4}$ & $82.0_{\pm 0.5}$ & $\mathbf{85.0}_{\pm 0.2}$ & -- \\
     & KEEP & selected & $88.1_{\pm 0.9}$ & $77.8_{\pm 2.7}$ & $81.7_{\pm 0.1}$ & $93.7_{\pm 0.9}$ & $82.4_{\pm 0.4}$ & $84.7_{\pm 0.5}$ & $-0.3$ \\
     & KEEP & static (restricted)& $88.7_{\pm 0.1}$ & $58.2_{\pm 1.3}$ & $69.1_{\pm 0.3}$ & $93.7_{\pm 2.0}$ & $80.6_{\pm 0.6}$ & $78.1_{\pm 0.4}$ & $-6.9$ \\
     & KEEP & static (un-restricted, masked) & $87.7_{\pm 0.4}$ & $72.9_{\pm 1.8}$ & $77.0_{\pm 0.1}$ & $93.3_{\pm 1.8}$ & $82.0_{\pm 1.2}$ & $82.6_{\pm 0.7}$ & $-2.4$ \\
    \cmidrule(lr){1-10}
    \multirow{4}{*}{75\%} & KEEP & dense & $81.9_{\pm 1.2}$ & $69.2_{\pm 1.8}$ & $77.1_{\pm 0.0}$ & $81.7_{\pm 1.1}$ & $73.5_{\pm 1.2}$ & $\mathbf{76.7}_{\pm 0.1}$ & -- \\
     & KEEP & selected & $78.8_{\pm 0.4}$ & $62.0_{\pm 3.4}$ & $73.8_{\pm 0.3}$ & $83.5_{\pm 3.2}$ & $71.0_{\pm 1.0}$ & $73.8_{\pm 0.4}$ & $-2.9$ \\
     & KEEP & static (restricted) & $74.6_{\pm 0.3}$ & $25.6_{\pm 5.3}$ & $28.6_{\pm 0.5}$ & $73.4_{\pm 1.8}$ & $60.9_{\pm 0.9}$ & $52.6_{\pm 1.3}$ & $-24.1$ \\
     & KEEP & static (un-restricted, masked) & $79.2_{\pm 2.9}$ & $49.5_{\pm 2.7}$ & $47.1_{\pm 0.3}$ & $77.0_{\pm 3.6}$ & $68.7_{\pm 1.3}$ & $64.3_{\pm 1.6}$ & $-12.4$ \\
    \bottomrule
  \end{tabular}
\end{table*}

To establish a point of comparison against DEP~\cite{koo2026dep}, we evaluated our method in the following settings: Instead of performing prefill with the full model, 1) we perform prefill only with the experts selected by our prediction, 2) we perform prefill on a static model with a restricted router, and 3) we perform prefill on a static model with an unrestricted router, where expert activations are tracked across all experts but execution remains masked to the static expert set. The difference between 2 and 3 is therefore whether the expert activation rate is tracked only within the pruned model or across all experts, while prefill itself remains restricted to the static expert set.

We test this on Qwen3.6-35B-A3B, and the results can be found in Table~\ref{tab:prefill}. We see that prefill with the full model performs best, while using the same set of experts for decoding and prefill comes second. Using a static model based on an expert ranking across all calibration data performs substantially worse, dropping by 12.4--24.1 points at 75\% sparsity. The model tracking activations across all experts performs better than the one tracking only the restricted set. However, both fall short of the setup where experts for prefill are selected based on the input. Therefore, we expect systems like DEP, where all experts for prefill and decoding are determined based on features of the prompt, to suffer from accuracy degradation compared to performing full prefill, even if they manage to predict the best experts. Furthermore, they face a harder prediction task, as they need to select experts for both prefill and decoding based only on the prompt, without observing the expert activation patterns during full prefill. This could explain their drops in accuracy, especially at higher sparsity levels.

\textbf{Full Prefill Economics.} However, the big dilemma is that only the two static pruning setups, with either a restricted router or an unrestricted router with masked execution, would really improve the memory situation by allowing us to run models that would otherwise exceed available memory. For the setup where we perform prefill and decoding on the same dynamically selected expert set, we would still need access to the full model to load the selected experts for each request. For models that exceed our memory capacity, this would again require reading experts from disk during prefill. Therefore, our prefill would still be bound by the speed of reading from disk, while introducing an additional accuracy trade-off. Consequently, we conclude that the accuracy degradation of pruned prefill does not justify its potential savings and choose to perform prefill on the full model.

\section{Additional End-To-End Results}
\label{sec:appendix_e2e}

\begin{figure*}[h]
    \centering
    \includegraphics[width=1\linewidth]{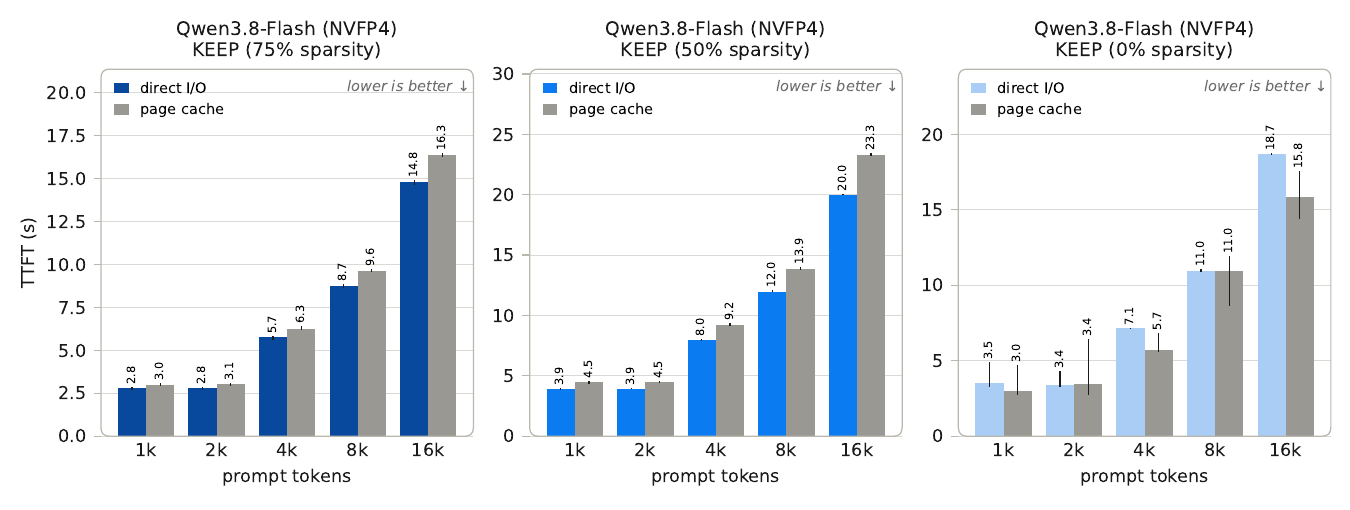}
    \caption{TTFT of Qwen3.8-Flash-Next (NVFP4) on the RTX 5090 with disk layers read through direct I/O or the page cache. Bars are medians over four runs after a warm-up.}
    \label{fig:page_misses_5090}
\end{figure*}

Using the results from our RTX 5090 run, we analyse how expensive our I/O ring is. We compare within the KEEP engine whether reading the disk layers with direct I/O or through the page cache is faster for prefill. 
In Figure~\ref{fig:page_misses_5090}, we see that for 50--75\% sparsity direct I/O was the right design choice, reducing TTFT by 8--16\%. Without pruning, the results are mixed and show no clear winner. We attribute this to free host memory: without a kept set, disk layers need no pinned pools, so the page cache can hold part of the disk layers between prefills.

We further examine the LRU cache miss rate in Table~\ref{tab:cache_misses}. At 75\% sparsity, the retained expert sets of both models fit almost entirely on the GPU, depending on the KV cache size, resulting in near-zero miss rates. For Qwen3.6, this already holds at 50\% sparsity. Interestingly, Qwen3.8-Flash-Next exhibits a slightly higher miss rate at 50\% sparsity than at 0\%. This is due to the different expert placement: at 0\% sparsity, layers whose experts reside on disk are executed on the CPU and therefore do not contribute to GPU cache misses. The remaining GPU-executed layers share the expert pool, leaving more cache slots available per layer.

\begin{table}[t]
  \centering
  \caption{Expert cache miss rates and routing behavior during decoding. 
  Outside-$S$ denotes the fraction of original router selections
  excluded by the retained expert set.}
  \label{tab:cache_misses}
  \small
  \setlength{\tabcolsep}{5pt}
  \begin{tabular}{@{}llrrr@{}}
    \toprule
    Model & Configuration
      & \shortstack{Miss rate\\(\%)}
      & \shortstack{Misses per\\layer-step}
      & \shortstack{Outside-$S$\\(\%)} \\
    \midrule
    \multirow{4}{*}{Qwen3.6-35B (FP8)}
      & KEEP (75\% sparsity)  & 0.002 & 0.0001 & 43.2 \\
      & KEEP (50\% sparsity)  & 0.001 & 0.0001 & 19.9 \\
      & KEEP (0\% sparsity) & 0.676 & 0.0543 & -- \\
      & FreeToken   & 0.676 & 0.0543 & -- \\
    \midrule
    \multirow{4}{*}{Qwen3.8-Flash (NVFP4)}
      & KEEP (75\% sparsity)  & 0.003 & 0.0003 & 33.8 \\
      & KEEP (50\% sparsity)  & 5.834 & 0.5834 & 12.9 \\
      & KEEP (0\% sparsity) & 4.985 & 0.4985 & -- \\
      & FreeToken (did not fit) & -- & -- & -- \\
    \bottomrule
  \end{tabular}
\end{table}

\begin{figure*}[h]
    \centering
    \includegraphics[width=1\linewidth]{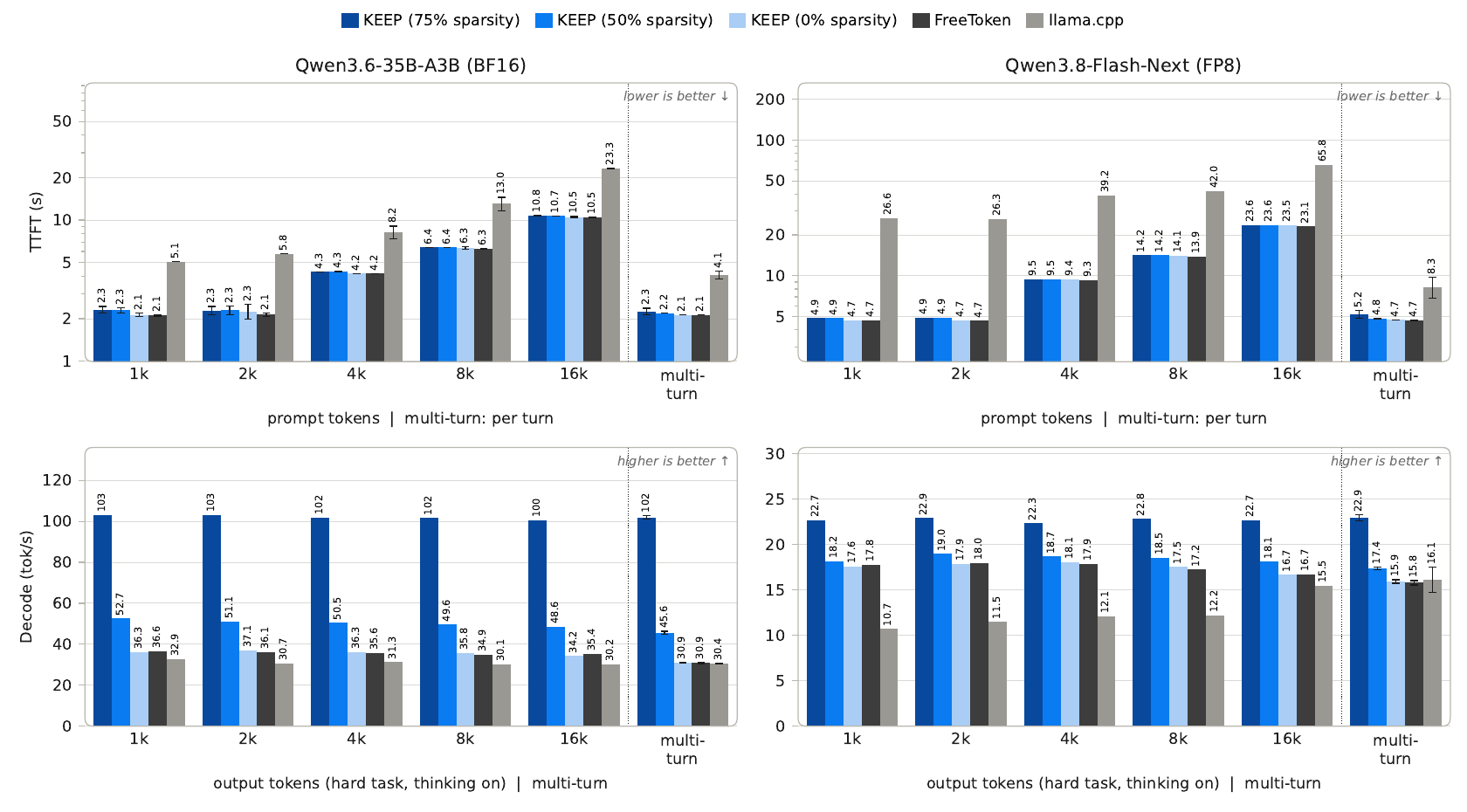}
    \caption{End-to-end evaluation on a server RTX 3090. TTFT by prompt length (top) and decode throughput by output length (bottom), and the multi-turn task as the last groups. Bars are medians over four runs after a warm-up. $\times$ means the model does not fit within our GPU+host memory.}
    \label{fig:latency_3090}
\end{figure*}

Lastly, we perform additional experiments on an RTX 3090 with enough host memory to fit both Qwen3.6-35B-A3B and Qwen3.8-Flash-Next in GPU+host memory. In Figure~\ref{fig:latency_3090}, we see that prefill overhead is negligible for both models and decreases with prompt length. Therefore, TTFT is on par with FreeToken or slightly worse, while always outperforming llama.cpp. Again, decoding benefits from sparsification, achieving up to 3$\times$ speedups over FreeToken and llama.cpp when the retained experts fit fully on GPU (Qwen3.6-35B-A3B), and up to 1.3$\times$ for Qwen3.8-Flash-Next, where only a small fraction of experts fits on GPU. Without sparsification, KEEP performs as well as FreeToken, showing no performance regression. 

Overall, these results show that KEEP improves decoding performance across different hardware and memory configurations while introducing negligible overhead.

\end{document}